\documentclass{article}

 \usepackage[preprint]{neurips_2026}

\usepackage[utf8]{inputenc} 
\usepackage[T1]{fontenc}    
\usepackage{hyperref}       
\usepackage{url}            
\usepackage{booktabs}       
\usepackage{amsfonts}       
\usepackage{nicefrac}       
\usepackage{microtype}      
\usepackage{xcolor}         

\usepackage{algorithm}
\usepackage{algpseudocode}
\usepackage{amsmath}
\usepackage{caption}
\usepackage{graphicx}
\usepackage{multirow}
\usepackage{subcaption}
\usepackage{tcolorbox}
\usepackage{wrapfig}

\usepackage[table]{xcolor}

\newcommand{\framework}{Region4Web}
\newcommand{\pipeline}{PageDigest}

\title{\framework: Rethinking Observation Space Granularity for Web Agents}

\author{%
  Donguk Kwon \\
  Yonsei University \\
  \texttt{donguk.kwon@yonsei.ac.kr} \\
  \And
  Dongha Lee \thanks{\; Corresponding author} \\
  Yonsei University \\
  \texttt{donalee@yonsei.ac.kr} \\
}

\begin{document}

\maketitle

\begin{abstract}
Web agents perceive web pages through an observation space, yet its granularity has remained an underexamined design choice.
Existing work treats observation at the same element-level granularity as the action space, leaving the page's functional organization implicit and forcing the agent to infer it from element-level signals at every step.
We argue observation should instead operate at the granularity of \textbf{functional regions}, parts of the page that each serve a distinct purpose.
We propose \textbf{\framework{}}, a framework that reorganizes the AXTree into functional regions through hierarchical decomposition and semantic abstraction, exposing the page's functional organization as the basis for page state understanding.
Moreover, we propose \textbf{\pipeline{}}, a web-specific inference pipeline that delivers this region-level observation to the actor agent as a compact per-page digest that persists across steps.
On the WebArena benchmark, \pipeline{} substantially reduces observation length while improving overall task success rate across diverse backbone large language models (LLMs) and established agent methods, regardless of backbone capacity.
These results show that operating at the granularity of functional regions delivers a more compact and informative basis for the actor agent than element-level processing alone.
Code is available at \small{\url{https://github.com/kwondu/region4web}}.
\end{abstract}

\section{Introduction}
Large language models (LLMs) have enabled autonomous agents capable of handling diverse real-world tasks in web environments~\citep{he2024webvoyager,logeswaran2026bookingarena,wu2025webwalker}.
At each step, a web agent perceives the current page state through an observation space and selects an action from an action space.
Prior work has concentrated on improving action selection, with task planning~\citep{guo2026webcogreasoner,huang2025r2d2,shinn2023reflexion}, element grounding~\citep{zheng2024seeact}, and model capability~\citep{qi2025webrl,wei2025webagentr1} all directed toward this goal.
Page state understanding, in contrast, has been addressed through filtering or truncating elements from the observation~\citep{kang2025acon,lee2025lcow,zhang2025prune4web}, which all operate at element-level granularity, leaving this design choice itself underexamined.

Existing work often represents the observation space at the same element-level granularity as the action space~\citep{schiepanski2025beyondpixels,yang2024agentoccam}, yet this granularity is not equally suited to both.
Element-level granularity is natural for the action space, where each action targets a specific element with a designated operation.
The observation space, however, serves a fundamentally different role of providing context for understanding the current page state, where context extends from individual elements to their relations.
We capture these relations through \textbf{functional regions}, defined as groups of elements whose relations support a shared purpose, such as site traversal or result narrowing.

\begin{figure}[t]
\centering
\includegraphics[width=\linewidth]{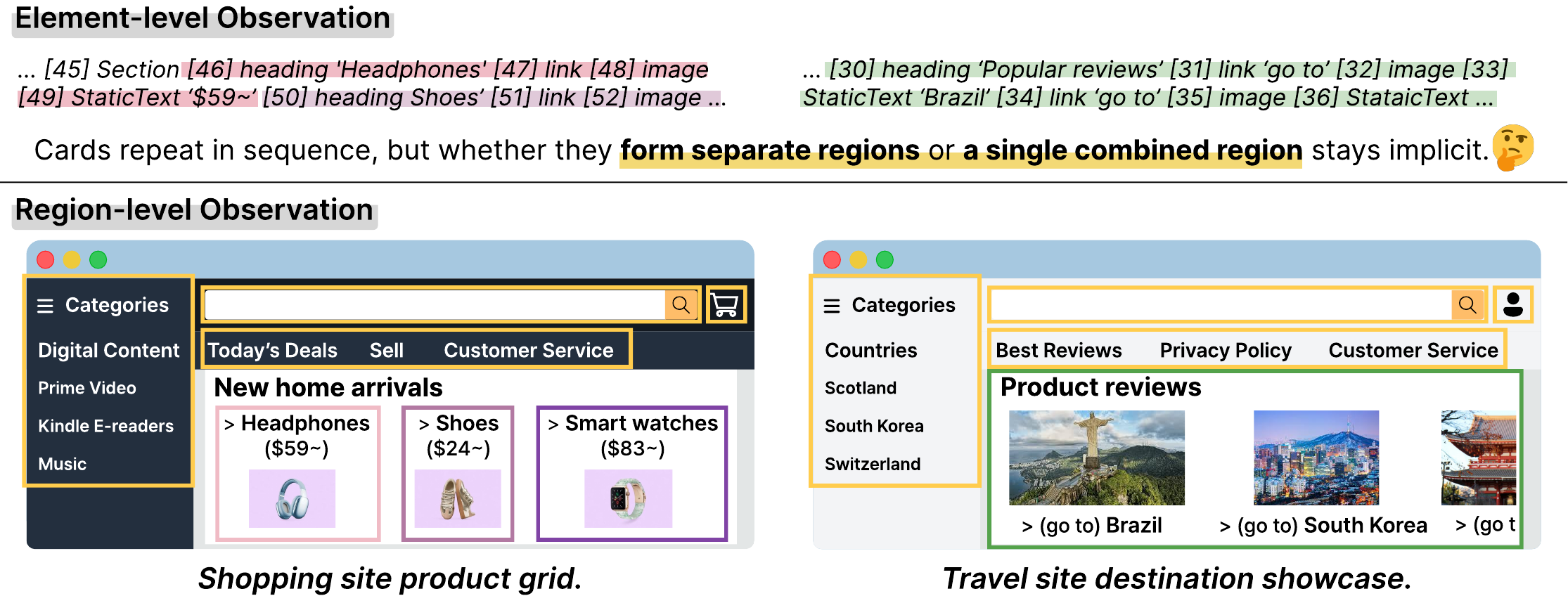}
\caption{Element-level and region-level observation of structurally similar card grids.
Region-level observation distinguishes a grid of product preview cards from a single destination showcase.}
\label{fig:region_observation}
\vspace{-1em}
\end{figure}

Decomposing pages into regions has been studied in human attention to spatially coherent areas~\citep{buscher2009surfing} and recent GUI web agents that segment screenshots into region partitions~\citep{fan2024tol,singh2025trishul}.
These approaches show that visual layout provides useful cues for grouping elements, often through spatial proximity such as bounding box overlap or layout adjacency.
However, spatial proximity does not entail shared functional purpose.
Such proximity cues may induce visual groupings, but do not specify whether they constitute functional observation units or what purpose they serve in the page state.
A similar implicitness appears in element-level observation~\citep{schiepanski2025beyondpixels,yang2024agentoccam,zhang2025prune4web}, where regions and their purposes are present only implicitly through individual elements and must be inferred by the agent.
Screenshot-based agents~\citep{he2024webvoyager,zheng2024seeact} provide layout cues that may make functional organization visually inferable, but they still require the agent to infer whether visually suggested groupings correspond to functional regions and what purpose they serve.
These limitations motivate region-level observation defined by shared functional purpose.
By identifying functional regions and abstracting each by its purpose, \framework{} makes page organization explicit before action selection, as shown in Figure~\ref{fig:region_observation}.

Constructing region-level observation is not straightforward.
Boundaries and purposes of functional regions are implicit in tree representations such as AXTree, where the hierarchy reflects markup nesting rather than how elements are organized.
Deriving them through rule-based decomposition is insufficient, as what each region is for varies with the page even for structurally repeated patterns.
A grid of structurally repeated cards, for example, forms independent regions when the cards are separate product previews, yet a single region when they collectively form a review showcase, as Figure~\ref{fig:region_observation} demonstrates.
Nor does existing research on web page structure resolve this, as web page segmentation~\citep{cai2003vips,gerber2025webclasseg,kiesel2020webseg} and content extraction~\citep{barbaresi2021trafilatura,liu2025dripper} methods target information retrieval or content analysis, not the functional organization that agent observation requires.
Its construction therefore demands learning how web pages are functionally organized across diverse page layouts.

We address this challenge with \textbf{\framework{}}, a framework that constructs region-level observation from the AXTree through two stages.
Hierarchical decomposition classifies each parent-child edge as merge or cut in a single bottom-up traversal, and the subtrees formed by merged edges constitute the functional regions of the page.
Semantic abstraction then interprets each region along two orthogonal dimensions, a purpose that identifies what the region is for and a state summary that captures its current actionable context.
Since both stages run at every page during agent execution, they are realized as small dedicated models.
The knowledge of how pages are functionally organized is implicit in the AXTree and cannot be derived by rule, so these models are trained on annotations from a proprietary LLM covering diverse real-world websites.

Moreover, deploying \framework{} in web environments requires keeping its region-level observation compact while preserving the page state understanding it supports, which motivates \textbf{\pipeline{}}, a web-specific inference pipeline that maintains a compact digest of the agent's observation across steps within each page.
Upon entering a new page, \pipeline{} selects task-relevant regions and exposes them as AXTree subtrees alongside the non-selected regions' abstractions, preserving element-level granularity for the action space within the page's structural information.
Within the same page, \pipeline{} tracks observation transitions across steps, rather than reconstructing the full observation at every step.
\pipeline{} shares the actor agent's backbone LLM and operates solely on the observation space, making it directly applicable to diverse web agents.

On the WebArena~\citep{zhou2024webarena} benchmark, \pipeline{} substantially reduces observation length across four backbone LLMs and two established agent methods, with the reduction holding consistently regardless of backbone capacity.
\pipeline{} improves overall task success rate across backbones, demonstrating that region-level observation strengthens page state understanding regardless of backbone capacity.
Ablations confirm that \framework{} and \pipeline{} make distinct contributions, with \framework{} alone supporting page state understanding while \pipeline{} delivers it compactly across steps.

Our contributions are summarized as follows.
\begin{itemize}
\item We propose \framework{}, a framework that reorganizes the AXTree into functional regions through hierarchical decomposition and semantic abstraction, exposing the page's functional organization as the basis for web agents' page state understanding.
\item We propose \pipeline{}, a web-specific inference pipeline that delivers each page's region-level observation to the actor agent as a compact digest that persists across steps, reducing observation length while preserving task success.
\item We evaluate \framework{} and \pipeline{} on the WebArena benchmark, where \pipeline{} substantially reduces observation length while improving overall task success rate, regardless of backbone capacity.
\end{itemize}

\section{Preliminary Analysis}
\label{sec:preliminary_analysis}
\begin{figure}[h]
\centering
\begin{subfigure}[b]{0.48\linewidth}
\centering
\includegraphics[width=\textwidth]{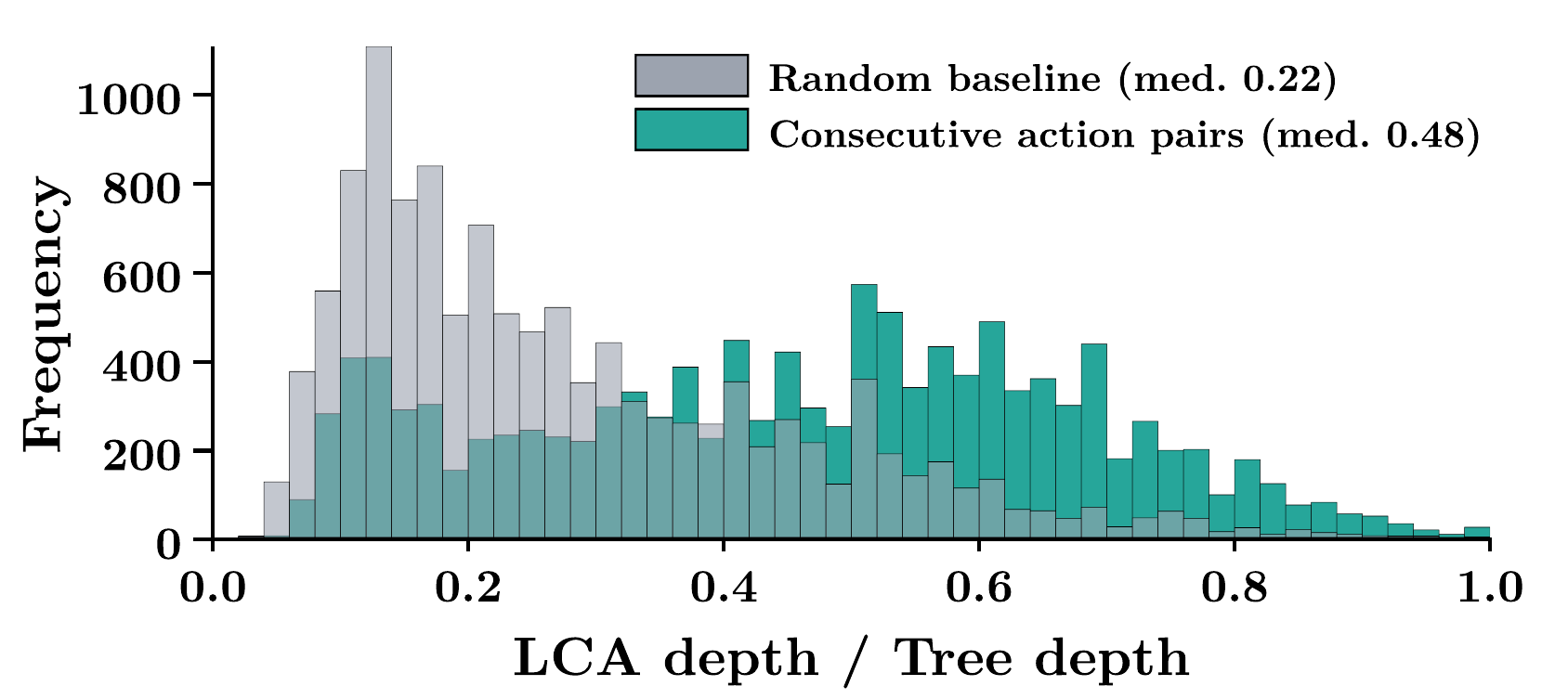}
\caption{Distribution of LCA depth ratio for consecutive action pairs against the random baseline.}
\label{fig:action_locality}
\end{subfigure}
\hfill
\begin{subfigure}[b]{0.48\linewidth}
\centering
\includegraphics[width=\textwidth]{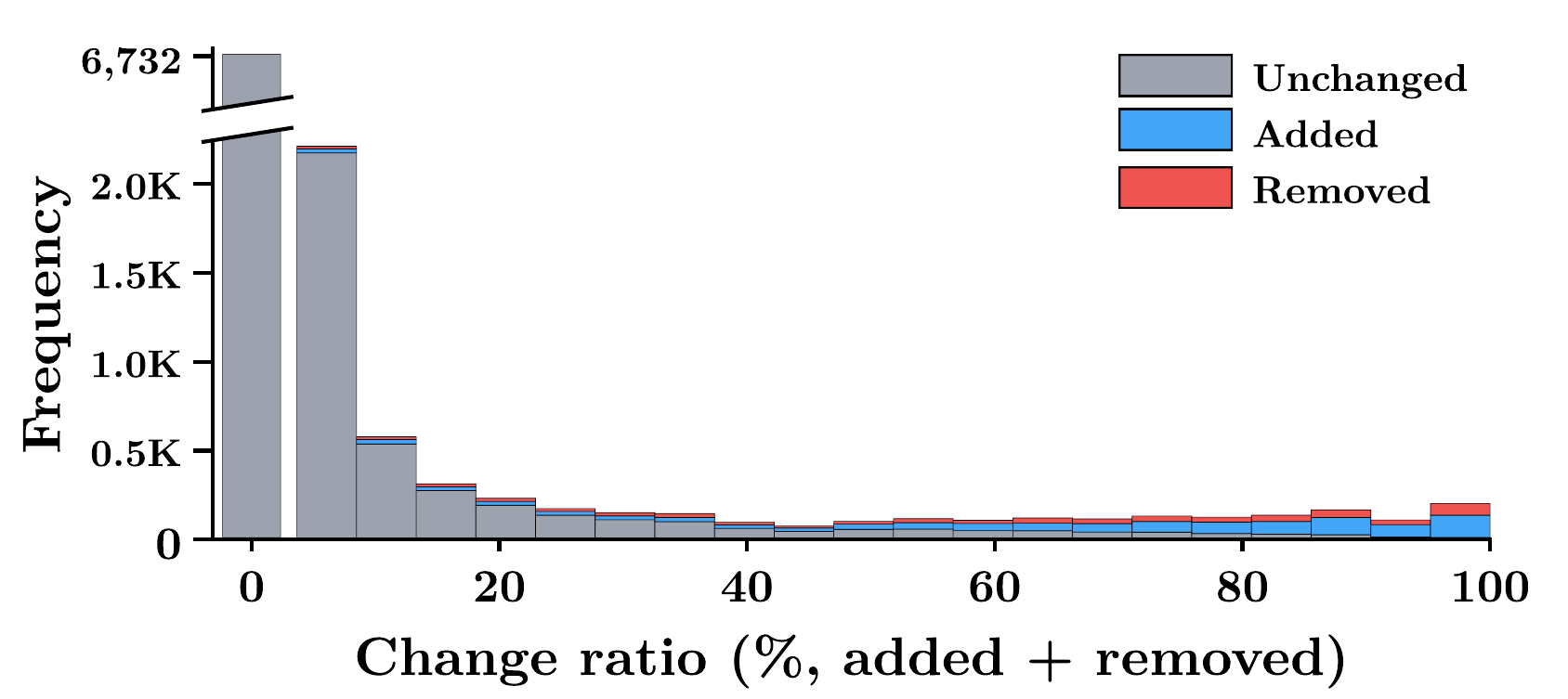}
\caption{Distribution of DOM change ratio across within-page steps, where 52.9\% exhibit zero change.}
\label{fig:observation_change}
\end{subfigure}
\end{figure}

We analyze action traces and observation transitions to inform two design questions about observation in web environments.
Section~\ref{sec:action_locality} examines whether the agent's actions are localized within the page structure during a task, motivating the unit at which observation should be constructed within a single step.
Section~\ref{sec:observation_change} examines how much the observation changes as the agent acts within a page, motivating the question of whether observation should be reconstructed at every step.

To answer these questions, we use the Mind2Web dataset~\citep{deng2023mind2web}, which provides 2,350 tasks with per-action ground-truth annotations across 137 real-world websites, with dataset selection criteria detailed in Appendix~\ref{apx:dataset_selection}.
Each page is represented as a DOM tree with an average of 2,473 nodes.
The dataset contains 15,394 consecutive action pairs, of which 12,009 (78.0\%) occur within the same page and the remaining 22.0\% involve page navigation that entirely replaces the observation.
Our analysis focuses on same-page pairs, where observation construction and update are at issue.

\subsection{Consecutive Actions Are Localized within Page Structure}
\label{sec:action_locality}

\paragraph{Only a negligible fraction of elements on a page are targeted during a task.}
While each page contains thousands of DOM nodes, the number of actions performed on it during a task has a median of 6 and a 90th percentile of 13.
Since each action targets exactly one element, the elements ever acted upon constitute a negligible fraction of the page.
The full page is thus dominated by elements irrelevant to the task, motivating selection of task-relevant content.

\paragraph{Consecutive actions are structurally co-located within the page.}
We measure the lowest common ancestor (LCA) depth ratio for consecutive action pairs, computed as the depth of the LCA of the two target elements divided by the maximum depth of the DOM tree.
A higher value indicates that the two elements are situated within a tighter subtree.
As Figure~\ref{fig:action_locality} shows, consecutive action pairs yield a median LCA depth ratio of 0.48, with 81.7\% exceeding the random baseline median of 0.22.
Consecutive actions thus concentrate within localized subtrees rather than spanning the page, indicating that the region serves as a natural unit for observation construction.

\subsection{Within-Page Observation Undergoes Marginal Change across Consecutive Steps}
\label{sec:observation_change}

For each step within a page, we measure the change ratio, the proportion of DOM elements added or removed by the action.
As Figure~\ref{fig:observation_change} shows, 52.9\% of steps exhibit zero change, and 74.4\% remain below 5\%.
Where changes occur, they reflect minor DOM modifications such as dropdown expansion or tooltip appearance.
Steps exceeding 90\% change account for only 2.5\%, attributed to client-side routing within single-page applications.
Reconstructing the full observation at every step is therefore unnecessary, and tracking only the incremental changes within each page can avoid this redundancy.

\section{\framework}
\label{sec:framework}
\begin{figure}[t]
\centering
\includegraphics[width=\linewidth]{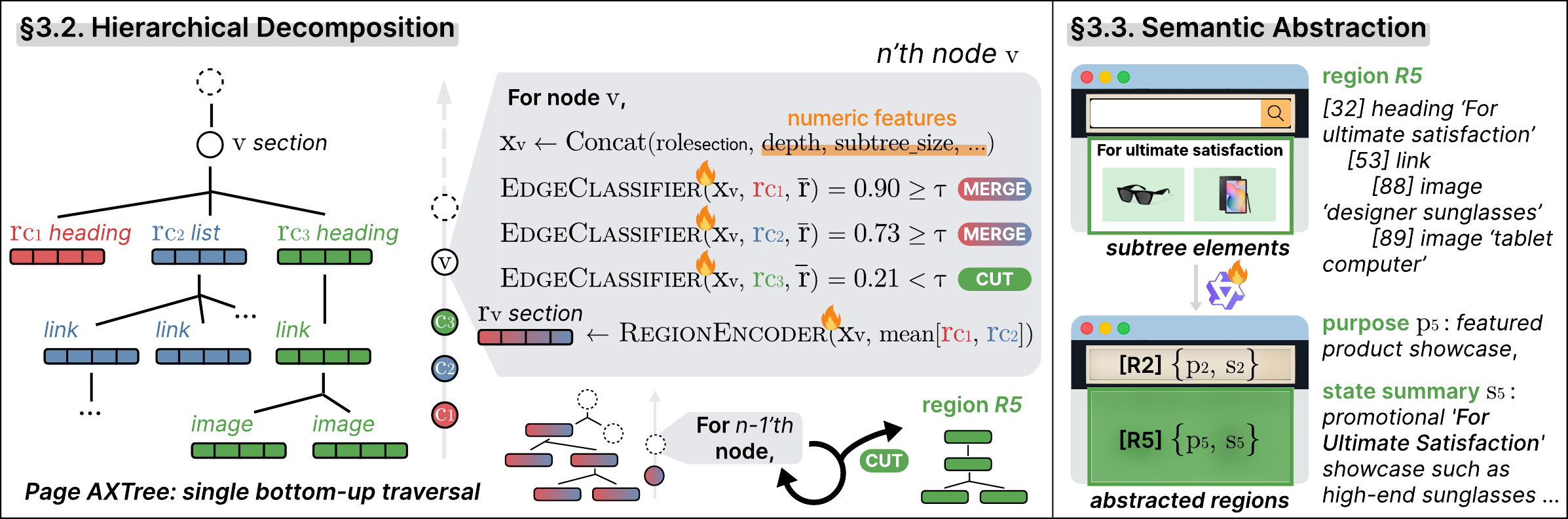}
\caption{Overview of \framework{} inference process.}
\vspace{-1em}
\end{figure}

Section~\ref{sec:action_locality} shows that regions are natural units for observation.
We propose \framework{}, a two-stage framework for constructing region-level observation from the AXTree of a web page.

\subsection{Problem Formulation}

At each step, a web agent perceives the current page state through an observation space and selects an action from an action space.
The observation can be represented as a tree $\mathcal{T} = (V, E)$, where each node $v \in V$ corresponds to an element on the page with attributes such as role, name, and value.
In the prevailing element-level approach, the agent operates over $V$ directly, leaving the page's functional organization implicit in $\mathcal{T}$.
Region-level observation makes this organization explicit through a partition $\mathcal{R} = \{R_1, \ldots, R_m\}$ of $V$ into functional regions, where each $R_i$ forms a subtree of $\mathcal{T}$.
Each region is associated with a purpose $p_i$ that identifies what the region is for and a state summary $s_i$ that captures its current actionable context.
\framework{} learns to produce both $\mathcal{R}$ and the associated $\{(p_i, s_i)\}$ from $\mathcal{T}$.

\subsection{Hierarchical Decomposition}

To construct region-level observation, $\mathcal{T}$ must be decomposed into a region partition $\mathcal{R}$.
We instantiate $\mathcal{T}$ as the page's AXTree, a browser-generated representation that encodes each element's accessibility semantics in a hierarchical structure.
Since each $R_i \in \mathcal{R}$ forms a subtree of $\mathcal{T}$, the partition is fully determined by classifying each edge in $E$ as merge or cut.
Removing the cut edges from $\mathcal{T}$ splits the tree into subtrees, each of which constitutes a region in $\mathcal{R}$.
Since the root has no parent edge to classify, its subtree constitutes the final region in $\mathcal{R}$ after the bottom-up traversal completes.

Decomposition determines region boundaries from structural cues alone, whereas semantic abstraction interprets each region's purpose and actionable state.
Each node $v$ is represented by a feature vector $\mathbf{x}_v$ that combines a learned role embedding with numeric features encoding the node's structural information in $\mathcal{T}$.
At each internal node $v$ with children $c_1, \ldots, c_k$ and their respective representations $\mathbf{r}_{c_1}, \ldots, \mathbf{r}_{c_k}$, an \textsc{EdgeClassifier} determines whether each child should be separated, using the sibling mean $\bar{\mathbf{r}} = \frac{1}{k}\sum_{j} \mathbf{r}_{c_j}$ as context,
\begin{equation}
\hat{y}_{v,c_i} = \textsc{EdgeClassifier}(\mathbf{x}_v,\; \mathbf{r}_{c_i},\; \bar{\mathbf{r}}).
\end{equation}
Edges with $\hat{y}_{v,c_i} \geq \tau$ are cut, while the remaining children $\mathcal{M}_v$ are merged into the parent's region.
\textsc{RegionEncoder} then computes the parent's representation from $\mathbf{x}_v$ and the merged children $\mathcal{M}_v$,

\begin{equation}
\mathbf{r}_v = \textsc{RegionEncoder}\!\left(\mathbf{x}_v,\; \frac{1}{|\mathcal{M}_v|}\sum_{c_j \in \mathcal{M}_v} \mathbf{r}_{c_j}\right),
\end{equation}
ensuring that the parent's representation reflects only the children that belong to its region.
For leaf nodes, since no children exist, $\mathcal{M}_v$ is empty and the aggregation term reduces to $\mathbf{0}$.

The entire procedure is carried out in a single bottom-up traversal, where each node's representation is computed only after all its children's boundary decisions are resolved, so that boundary decisions propagate upward through the hierarchy without requiring an additional pass.
The full procedure is detailed in Algorithm~\ref{alg:decomposition}.

\subsection{Semantic Abstraction}

The region partition $\mathcal{R}$ determines which elements belong together, but the semantic meaning of each region remains implicit in its subtree.
A fine-tuned language model receives the preprocessed AXTree subtree of each region and produces a purpose $p_i$ and a state summary $s_i$, which address two orthogonal dimensions.
Purpose captures what the region is for, serving as the basis for identifying each region.
State summary interprets the region's current actionable context, conveying what information and actionable elements are available within it.

\subsection{Training}

Since both stages run on every page during agent execution, they are realized as small dedicated models, a decomposition model for structural boundary decisions and a small language model for per-region abstraction.
The knowledge of how web pages are functionally organized is implicit in $\mathcal{T}$ and cannot be derived by rule, so these models are trained on annotations from a proprietary LLM.
We employ \texttt{gpt-5-mini-2025-08-27}~\citep{openai2025gpt5} as the annotator to construct the training dataset.
The raw AXTree is preprocessed into a textual form that retains the elements an agent can perceive and act on along with the structural grouping among them.
Since \framework{} operates sequentially, with decomposition producing $\mathcal{R}$ that abstraction then interprets, the training dataset should be constructed to follow this same dependency.

\paragraph{Dataset Construction.}
Source pages are collected from 500 real-world websites sampled from the Tranco top-1M ranking list,~\footnote{Tranco top-1M ranking list snapshot from April 1, 2026. \url{https://tranco-list.eu/list/QWQ94/1000000}} a research-oriented ranking of most popular websites.
These websites span 10 domain categories (e.g., Technology \& Computing, Shopping) derived from the IAB Content Taxonomy,~\footnote{IAB Content Taxonomy 3.1. \url{https://iabtechlab.com/standards/content-taxonomy}} a standard classification of web content, for their relevance to web agent tasks.
For each website, up to 100 page URLs are sampled using a score computed from sitemap metadata, yielding 21,974 pages from 253 websites whose AXTrees are successfully extracted.
The annotator then processes each page in three steps.
It first decomposes the AXTree into a region partition, then verifies the partition to identify incorrectly formed regions, and finally produces a purpose and a state summary for each verified region.
Only pages whose partitions contain no invalid region are retained, yielding 2,052 pages and 45,147 regions.
Pages excluded by this filter are dominated by real-world website noise that prevents coherent region organization rather than by annotator capacity, so the retained pages carry reliable annotations.

\paragraph{Decomposition training.}
The verified region partitions are converted into binary edge labels over $E$, where each edge is labeled as cut if its parent and child belong to different regions and as merge otherwise.
The model is trained with teacher forcing, where ground-truth labels determine the cut and merge decisions during the bottom-up traversal so that each node's representation is computed from correctly partitioned children.
Since merge edges vastly outnumber cut edges, focal loss with $\alpha = 0.75$ and $\gamma = 2.0$ is applied to address the class imbalance~\citep{lin2017focal,yihong2025cilg}.

\paragraph{Abstraction training.}
Qwen3-0.6B~\citep{yang2025qwen3} is fine-tuned on the 45,147 region annotations from the verified pages, with each example pairing a region's preprocessed subtree as input with the corresponding purpose and state summary as output.
A small model is chosen so that abstraction can be invoked once per region without dominating inference latency.

\noindent Further details on AXTree preprocessing, dataset construction, and \framework{} implementation are provided in Appendices~\ref{apx:axtree_processing}, \ref{apx:dataset_construction}, and \ref{apx:framework_implementation}, respectively.

\section{\pipeline}
\label{sec:pipeline}
\begin{figure}[t]
\centering
\includegraphics[width=\linewidth]{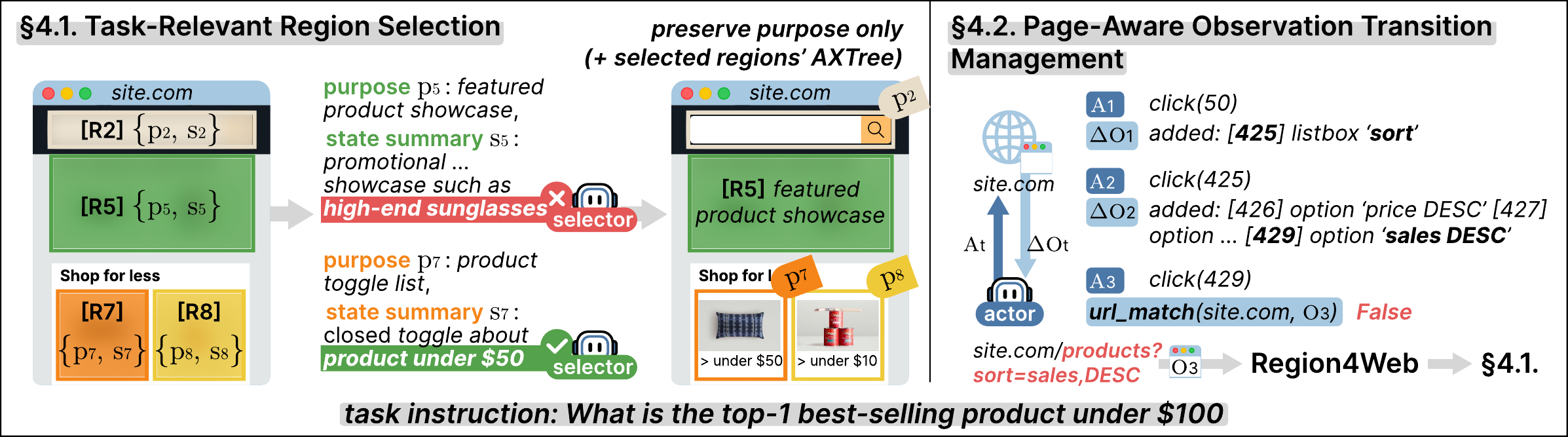}
\caption{Overview of \pipeline{}.}
\vspace{-1em}
\end{figure}

\framework{} produces region-level observation for a given page, but deploying it in web environments requires focusing the observation on what is task-relevant and tracking how pages change as the agent acts.
We propose \pipeline{}, a web-specific inference pipeline that constructs a page digest upon entering a new page through region selection, retains it within the page, and updates it through observation transition tracking across steps.

\subsection{Task-Relevant Region Selection}
\label{sec:relevant_selection}

Upon entering a new page, \framework{} produces the region partition $\mathcal{R}$ and the associated $\{(p_i, s_i)\}$ for the page.
The actor agent's backbone LLM takes the abstractions $\{(p_i, s_i)\}$ together with the task instruction and the action history taken so far, and selects the task-relevant regions.
The abstractions specify each region individually and collectively convey the page's overall functional structure and current state, from which the model infers where the task currently stands and what is required next.
Selected regions are exposed to the actor agent as their AXTree subtrees with their purposes, preserving element-level granularity for the action space, while non-selected regions are represented by their purposes alone, retaining the page's overall structural information.

\subsection{Page-Aware Observation Transition Management}
\label{sec:transition_management}

Section~\ref{sec:observation_change} shows that within-page observation changes only marginally between consecutive steps.
\pipeline{} therefore tracks observation transitions across steps, rather than reinvoking \framework{} at each step.
During transition management, only the region purposes are referenced, since each purpose describes what its region is for and remains stable across steps within the page.
State summaries, in contrast, describe each region's current actionable context, making them useful for region selection at page entry but less suitable for page-aware observation transition management.

At each step, observation transitions are identified by comparing the current AXTree against its state at page entry, yielding added, removed, and modified nodes.
Removed and modified nodes update the AXTree constructed upon entering the page by deleting nodes or changing their values under the existing region purposes.
Added nodes, in contrast, are listed as a separate group that retains the structural grouping among them, since merging them into existing regions could shift those regions' purposes.
The actor agent thus receives the current observation across steps within the page, preserving continuity of the page state.
When the agent navigates to a new page, signaled by a URL change, \framework{} is invoked to produce the new page's $\mathcal{R}$ and $\{(p_i, s_i)\}$, and region selection proceeds.

\pipeline{} shares the actor agent's backbone LLM and operates solely on the observation space, requiring no additional model and leaving the actor agent's policy unmodified, making it directly applicable to diverse web agents.
Moreover, since region selection is performed by the actor agent's backbone and depends on its capability, the actor agent is given additional \texttt{view\_all} action that reveals all regions in their full AXTree subtree form for the remainder of the page, providing a fallback when the selected regions are insufficient.

\section{Experiments}
\label{sec:experiments}
\subsection{Experimental Setup}
\label{sec:experimental_setup}

\paragraph{Evaluation benchmark.}
We evaluate on WebArena~\citep{zhou2024webarena}, a comprehensive web agent benchmark that spans five distinct domains, namely e-commerce, social forum, collaborative development, content management, and map services.
Its 812 long-horizon tasks, each allowing up to 30 steps, cover diverse interaction patterns.
Since the original evaluator relies on \texttt{gpt-4-1106-preview} for fuzzy answer matching, which has since been deprecated, we replace it with GPT-4o.

\paragraph{Actor agents.}
We evaluate across diverse backbone LLMs and actor agent methods to verify that \pipeline{} consistently reduces observation length regardless of the model's capability or the agent's design while preserving the performance.
The backbone LLMs span two proprietary and two open-source LLMs, namely GPT-5.1~\citep{openai2025gpt5}, Gemini 3.1 Flash-Lite~\citep{google2026gemini31flash}, Deepseek-V3.2~\citep{deepseekai202deepseekv32}, and Qwen3.5-27B~\citep{yang2025qwen3}.
Each backbone selects the next action given the interaction history and current observation at each step~\citep{yao2022react}.
We further evaluate on two established agent methods widely adopted in WebArena evaluation.
SteP~\citep{sodhi2024step} dynamically composes human-designed LLM policies tailored to WebArena tasks through a stack-based Markov decision process.
AgentOccam~\citep{yang2024agentoccam} refines the observation and action spaces to align them with the underlying LLM's pretrained capabilities.
Since AgentOccam runs with its own space alignment, in the \pipeline{} configuration, we replace its alignment with \pipeline{} while retaining the action space alignment, isolating the effect of region-level observation.
We evaluate SteP and AgentOccam with GPT-4o as the backbone, matching the GPT-4 family under which both methods were originally developed.

\paragraph{Implementation details.}
All experiments are conducted in the BrowserGym environment, with Map domain tasks routed to the live OpenStreetMap service~\footnote{OpenStreetMap service. \url{https://www.openstreetmap.org}} following~\citep{chae2025wma,zhang2026planmcts}.
For reproducibility, open-source models are run at temperature 0 with thinking mode disabled where applicable, while proprietary models retain their default configuration.
We define observation length as the token count of the observation provided at the agent at each step.
All token counts reported in the experiments are measured under the OpenAI \texttt{o200k\_base} tokenizer.
All prompts are provided in Appendix~\ref{apx:prompts}.

\begin{table}[t]
\centering
\small
\caption{WebArena success rate (\%) across domains, with the average observation token length per step reported in the Obs. length column. Each actor agent is reported with and without \pipeline{}.}
\label{tab:main_results}
\begin{tabular}{lcccccc|c}
\toprule
Actor agent & Shopping & CMS & Reddit & GitLab & Map & Overall & Obs. length ($\Delta$) \\
\midrule
\multicolumn{7}{l}{\textit{Proprietary LLMs}} & \\
\midrule
GPT-5.1
& 39.0 & \textbf{57.1} & \textbf{65.1} & 46.1 & 24.8 & 45.2 & 6,116 \\
\rowcolor{gray!25} \quad + \pipeline{}
& \textbf{41.1} & 54.5 & 60.5 & \textbf{50.5} & \textbf{28.4} & \textbf{47.5} & \textbf{4,302 (-30\%}) \\
Gemini 3.1 Flash-Lite
& 33.3 & 41.2 & 54.0 & \textbf{44.7} & 22.9 & 39.4 & 6,705 \\
\rowcolor{gray!25} \quad + \pipeline{}
& \textbf{34.9} & \textbf{44.5} & \textbf{59.3} & 42.6 & \textbf{28.4} & \textbf{41.6} & \textbf{3,207 (-52\%}) \\
\midrule
\multicolumn{7}{l}{\textit{Open-source LLMs}} & \\
\midrule
DeepSeek-V3.2
& 30.7 & 51.1 & 58.5 & 40.4 & \textbf{21.1} & 40.4 & 7,158 \\
\rowcolor{gray!25} \quad + \pipeline{}
& \textbf{33.3} & \textbf{53.7} & \textbf{61.1} & \textbf{46.5} & \textbf{21.1} & \textbf{43.4} & \textbf{4,521 (-37\%)} \\
Qwen3.5-27B
& 22.9 & \textbf{53.8} & 58.8 & \textbf{35.3} & \textbf{22.9} & 38.2 & 5,767 \\
\rowcolor{gray!25} \quad + \pipeline{}
& \textbf{38.6} & 46.2 & \textbf{60.4} & 35.2 & 18.9 & \textbf{39.9} & \textbf{2,654 (-54\%)} \\
\midrule
\multicolumn{7}{l}{\textit{Web agent methods} (backbone: GPT-4o)} & \\
\midrule
SteP~\citep{sodhi2024step}
& 32.6 & \textbf{45.7} & \textbf{71.4} & 46.9 & 12.8 & \textbf{39.5} & 7,136 \\
\rowcolor{gray!25} \quad + \pipeline{}
& \textbf{35.8} & 37.1 & 63.2 & \textbf{50.0} & \textbf{15.4} & 38.7 & \textbf{3,693 (-50\%)} \\
AgentOccam~\citep{yang2024agentoccam}
& 26.7 & 36.3 & \textbf{73.7} & \textbf{66.7} & 11.5 & 40.6 & 4,025 \\
\rowcolor{gray!25} \quad + \pipeline{}
& \textbf{35.6} & \textbf{40.0} & 68.4 & 50.0 & \textbf{23.1} & \textbf{41.3} & \textbf{3,365 (-16\%)} \\
\bottomrule
\end{tabular}
\vspace{-1.5em}
\end{table}

\subsection{Main Results}

\paragraph{\pipeline{} improves overall task success rate while reducing observation length, regardless of backbone capacity.}
Across the four backbones in Table~\ref{tab:main_results}, \pipeline{} reduces observation length by 43\% on average, from 6,437 to 3,671 tokens, and improves task success rate by 2.3\%p on average.
The improvement holds across backbones of varying capacity, suggesting that region-level observation provides a complementary signal for page state understanding that benefits backbones independent of their strength.

\paragraph{\pipeline{} extends to established agent methods through the observation space.}
Applying \pipeline{} to SteP and AgentOccam reduces observation length by 50\% and 16\% with comparable task success rate.
For AgentOccam, replacing its observation space alignment with \pipeline{} yields comparable performance, showing region-level observation can replace element-level alignment for action selection.
Since \pipeline{} operates solely on the observation space, sharing the actor's backbone, it applies to diverse web agents, with task success scaling with backbone capacity.

\subsection{Further Analysis}

We further analyze the contributions of \framework{} and \pipeline{} using GPT-5.1, with case studies of \framework{}'s decomposition and abstraction in Appendix~\ref{apx:case_study}.

\begin{wraptable}{r}{0.5\linewidth}
\vspace{-1em}
\centering
\caption{Ablation on WebArena-Lite. The Obs. length column matches Table~\ref{tab:main_results}.}
\label{tab:ablation}
\small
\begin{tabular}{lcc}
\toprule
Configuration & SR (\%) & Obs. length \\
\midrule
GPT-5.1
& 48.5 & 5,410 \\
\quad + \framework{}
& \underline{50.3} & 5,922 \\
\quad + Self-ctx~\citep{lee2025lcow} + §~\ref{sec:transition_management}
& 46.1 & \underline{4,013} \\
\midrule
\rowcolor{gray!25} \quad + \pipeline{}
& \textbf{53.9} & \textbf{3,814} \\
\bottomrule
\end{tabular}
\end{wraptable}

\paragraph{\framework{} improves page state understanding, while \pipeline{} keeps it compact across steps.}
For the ablation study, we use WebArena-Lite~\citep{lee2025lcow,liu2024visualagentbench}, a 165-task subset of WebArena.
Table~\ref{tab:ablation} compares the backbone, \framework{} alone, an element-level variant of \pipeline{} that omits \framework{} and replaces the region selection stage with self-contextualization as in LCoW~\citep{lee2025lcow}, and \pipeline{}.
\framework{} alone improves task success rate from 48.5\% to 50.3\% with comparable observation length, while the element-level variant lowers it to 46.1\%, showing that region-level observation supports the actor agent where element-level processing instead hinders it.
\pipeline{} reduces observation length by 30\%, comparable to the element-level variant's 26\% reduction, while still achieving the highest task success rate among the configurations, improving over the backbone by 5.4\%p.
Together, these results show that page state understanding and compact persistence are complementary, with \framework{} preserving functional regions for page state understanding and \pipeline{} keeping them compact across steps.

\begin{figure}[h]
\centering
\begin{subfigure}[b]{0.48\linewidth}
\centering
\includegraphics[width=\textwidth]{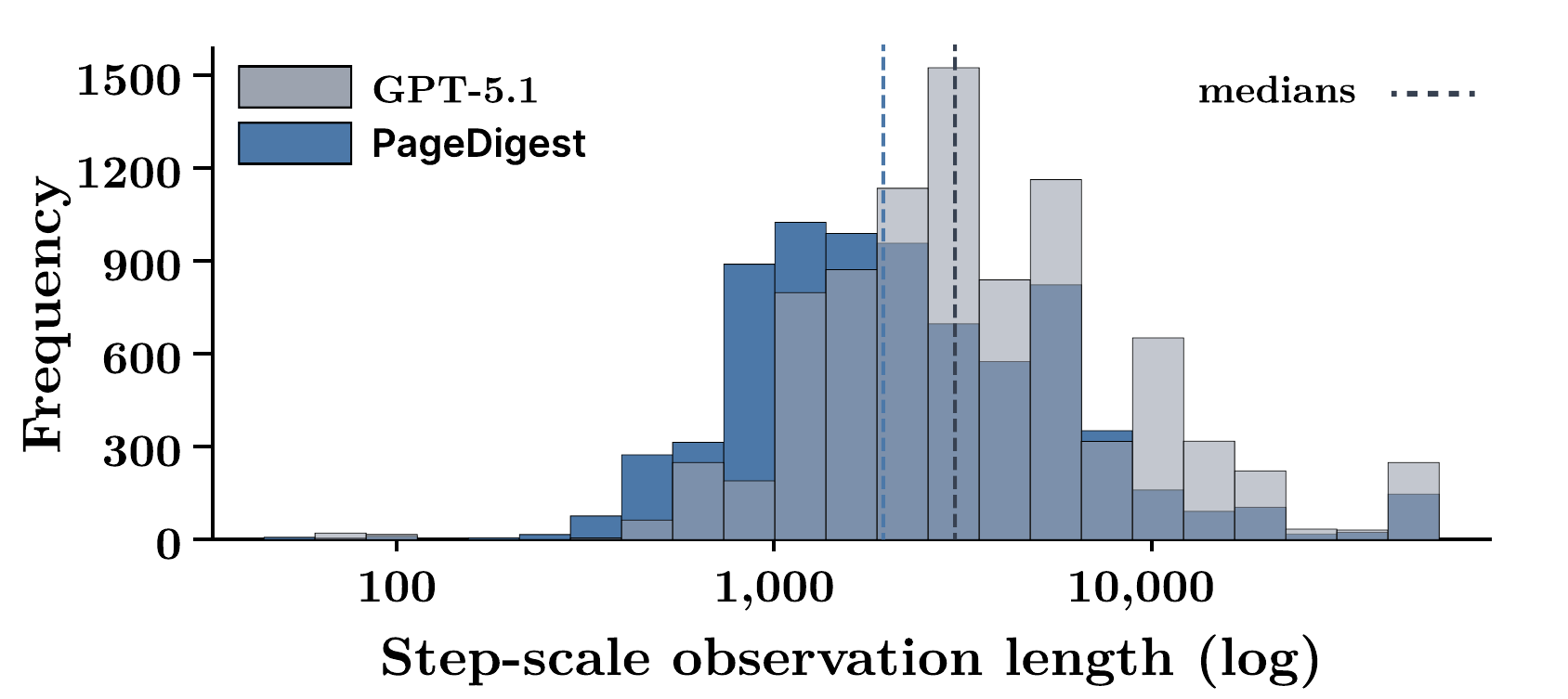}
\caption{Distribution of step-scale observation length.}
\label{fig:step_length}
\end{subfigure}
\hfill
\begin{subfigure}[b]{0.48\linewidth}
\centering
\includegraphics[width=\textwidth]{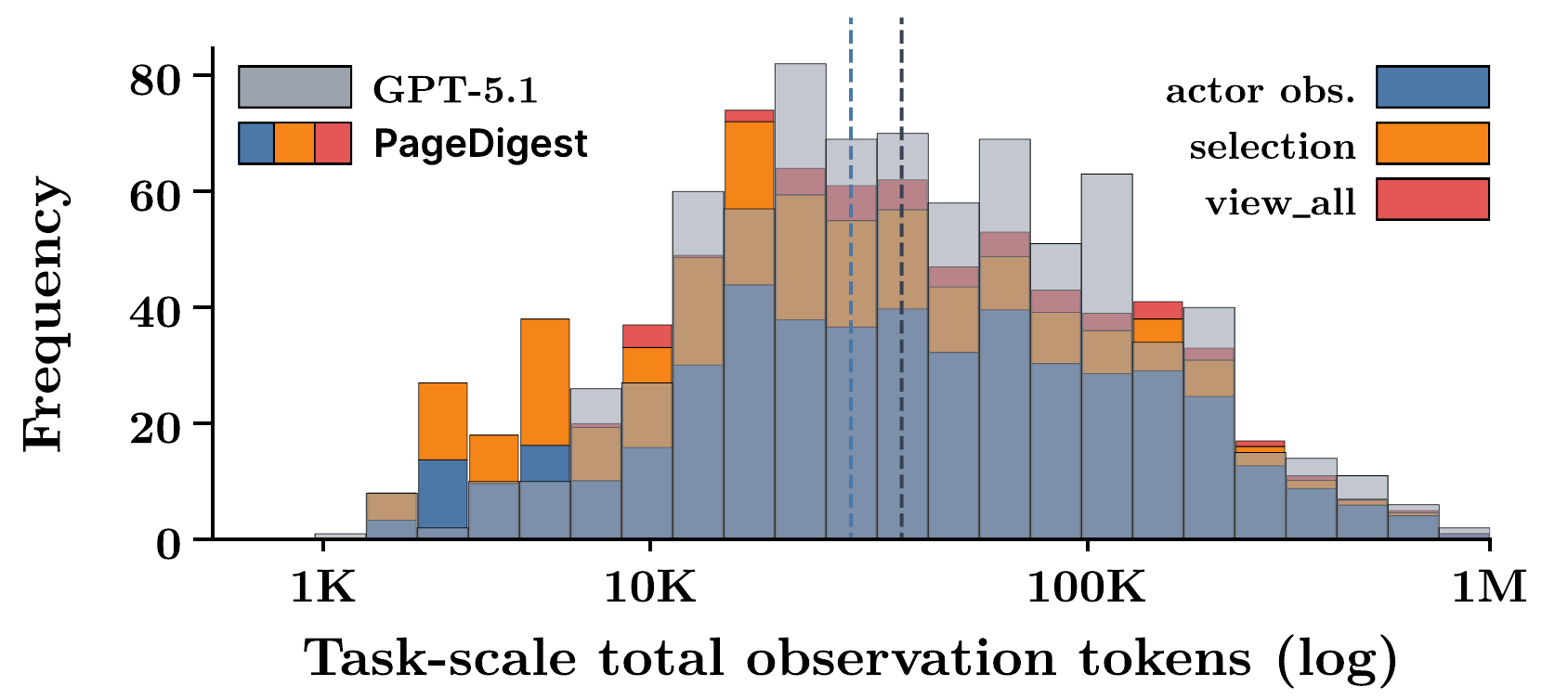}
\caption{Distribution of task-scale total observation tokens.}
\label{fig:task_length}
\end{subfigure}
\vspace{-0.5em}
\end{figure}

\paragraph{\pipeline{} preserves its step-scale reduction at the task-scale despite the auxiliary inference it adds.}
As Figure~\ref{fig:step_length} shows, \pipeline{} reduces the median observation length by 33\%, from 3,077 to 2,066 tokens, and as Figure~\ref{fig:task_length} shows, the median cumulative observation across a task drops by 25\%, from 26,707 to 19,944 tokens.
The task total comprises more than the actor observations alone, since region selection inputs are added at each entry into a new page, and \texttt{view\_all} expansions are added whenever the fallback is triggered.
Decomposing the task total, the actor observation accounts for 73.9\%, region selection for 19.5\%, and \texttt{view\_all} for 6.6\%.
Region selection stage stays cheap since it operates over the region-level abstractions $\{(p_i, s_i)\}$ rather than the element-level AXTree, even when invoked 4.8 times on average across a task, and \texttt{view\_all} is invoked sparingly in only 38.1\% of tasks, with an average of 0.64 calls across a task.
The auxiliary overhead therefore stays bounded by page entries, and the step-scale compactness carries through to the task-scale.

\paragraph{\pipeline{}'s failures lie largely outside its own design.}
We randomly sample 50 failed task trajectories under \pipeline{} on WebArena, 10 from each domain, and trace the failure to its

\begin{wrapfigure}{r}{0.46\linewidth}
\centering
\includegraphics[width=\linewidth]{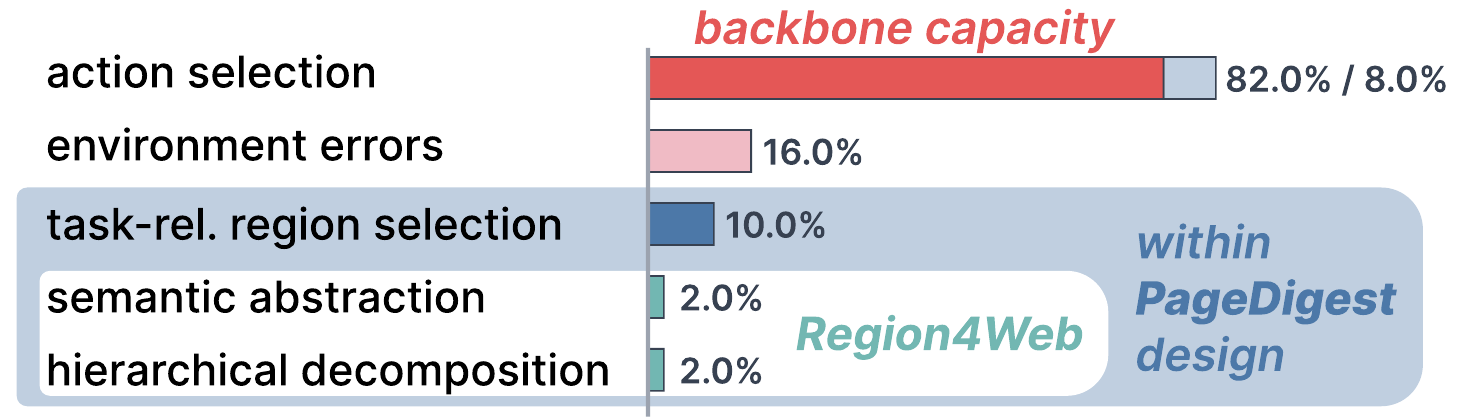}
\caption{Failure mode distribution under \pipeline{} on WebArena.}
\label{fig:failure_mode}
\vspace{-1em}
\end{wrapfigure}

\noindent triggering step and label every \pipeline{} stage, as shown in Figure~\ref{fig:failure_mode}.
Decomposition and abstraction errors (each 2.0\%) together account for only a small fraction, indicating that \framework{} reliably decomposes and abstracts regions on most pages.
Selection errors (10.0\%) reflect the backbone LLM missing task-relevant regions despite \framework{}'s informative abstractions.
Transition management introduces no errors, as it deterministically compares the current AXTree against its state at page entry.
Actor-side failures in the backbone's action selection account for 90.0\%, with environment errors outside the pipeline adding 16.0\%.
\pipeline{} thus operates as designed, with 82.0\% backbone capacity dominating 8.0\% \pipeline{} regression under multi-cause attribution.

\section{Related Work}
\paragraph{Web Page Structure Understanding}
has been studied for information retrieval and content analysis, treating web pages as content to be processed rather than as observation for agents.
Web page segmentation partitions pages into visually or structurally coherent blocks, exemplified by VIPS~\citep{cai2003vips}.
Subsequent work has focused on evaluation methodology~\citep{kiesel2020webseg,kiesel2021comparison} and macro-structural labels such as header, main content, and footer~\citep{gerber2025webclasseg}.
Content extraction separates main content from surrounding noise through rule-based heuristics~\citep{barbaresi2021trafilatura} or language models~\citep{chen2025indexlm,liu2025dripper,wang2025readerlmv2}.
In contrast, our \framework{} constructs region-level observation for web agents by decomposing the page into functional regions and making each region's purpose explicit for action selection.

\paragraph{Observation Processing in Web Agents}
has explored strategies to reduce observation length while preserving task-relevant information.
A dominant line focuses on element selection~\citep{moskaleva2025focusagent}, where Prune4Web~\citep{zhang2025prune4web} filters elements via LLM-generated keyword matching programs, and LCoW~\citep{lee2025lcow} trains a contextualization module that extracts task-relevant elements and annotates them contextually.
Orthogonal to selection, Beyond Pixels~\citep{schiepanski2025beyondpixels} downsamples the DOM tree while preserving its hierarchical structure.
AgentOccam~\citep{yang2024agentoccam} reformulates elements into markdown and identifies pivotal nodes to retain across steps.
Multimodal web agents use screenshots as additional input~\citep{guo2026webcogreasoner,he2024webvoyager,zheng2024seeact}, while recent GUI agents introduce visually decomposed region structures~\citep{fan2024tol,singh2025trishul}.
These approaches provide visual or layout cues for understanding the page state, but they do not define observation units by shared functional purpose, leaving which elements form functional regions and what purposes those regions serve implicit.
Our work treats observation granularity as a design choice, shifting from element-level to region-level observation and deploying it through a web-specific inference pipeline.

\paragraph{Tree-Structured Representation Learning}
has been studied across domains, from syntactic parse trees in natural language processing~\citep{tai2015treelstm}, to abstract syntax trees in source code analysis~\citep{mou2016tbcnn,wang2021modulartree,zhang2019astnn}, to DOM trees in web page understanding~\citep{wang2022webformer,yeoh2022grownup}.
These methods typically compute representations over a fixed tree structure and use the resulting node or tree representations for downstream prediction.
In this design, the tree structure is given in advance, and representation learning does not change which children belong to each parent.
Region partitioning breaks this independence, as boundary decisions directly alter the set of children a parent must represent.
This boundary-representation dependency motivates the joint computation in a single bottom-up traversal that \framework{} adopts.

\section{Conclusion}
We presented \framework{} and \pipeline{}, addressing observation granularity as an underexamined design choice for web agents.
\framework{} reorganizes the AXTree into functional regions to support the actor agent's page state understanding, and \pipeline{} delivers this region-level observation as a compact digest that persists across steps.
On the WebArena benchmark, \pipeline{} substantially reduces observation length while improving overall task success rate across diverse backbone LLMs and established agent methods, demonstrating that region-level observation can provide a more compact and informative basis for web agent decision making than element-level processing.
These results show that observation granularity directly affects web agent efficiency. By separating observation design from model capability and action policy, our work opens a path toward more efficient web agents by rethinking the granularity at which pages are observed.

\small
\bibliography{reference}
\bibliographystyle{plain}








\appendix

\section{Limitations and Future Work}
\label{apx:limitations}

\framework{} operates over the AXTree, and its decomposition and abstraction quality therefore depend on how completely each page exposes its accessibility semantics, with pages that render through canvas or rely on non-semantic markup providing weaker structural cues for boundary classification.
Our evaluation focuses on WebArena, whose consistent AXTree fidelity supports the controlled comparisons our experiments require.
Broader validation on real-world web environments, where AXTree fidelity and page complexity vary across sites, complements these results.
Future work includes broadening evaluation to live web environments such as Online-Mind2Web~\citep{xue2025onlinemind2web}, and applying region-level granularity to screenshot-based agents, since organizing observation by shared functional purpose generalizes across modalities.

\section{Broader Impacts}
\label{apx:broader_impacts}

\framework{} and \pipeline{} reduce the observation length required for web agent operation, which lowers inference cost and broadens access to web agent technology in resource-constrained settings.
The same efficiency gains can also lower the barrier to misuse such as large-scale scraping or automated abuse of online services, where mitigation lies at the deployment level through controls such as rate limiting and access policies.
Our training data is constructed from publicly accessible pages on Tranco-listed websites and contains no personal or sensitive information, limiting privacy concerns from the released artifacts.

\section{Dataset Selection for Preliminary Analysis}
\label{apx:dataset_selection}

Our preliminary analysis requires a web agent benchmark that provides per-action ground-truth annotations across diverse real-world web pages, so that consecutive action targets can be identified and localized within the page structure.
Mind2Web~\citep{deng2023mind2web} is well suited for this purpose.
It provides 2,350 tasks across 137 websites spanning 31 domains, where each action step is grounded in the DOM snapshot of the page at that step.
Since our analysis targets structural properties within individual snapshots, the static nature of these representations does not affect the validity of the measurements.

Other web agent benchmarks do not meet these requirements.
MiniWoB++~\citep{liu2018miniwob} consists of atomic-level tasks in synthetic web environments that do not reflect the structural complexity of real-world pages.
Mind2Web-Live~\citep{pan2024webcanvas} provides tasks on live websites, but its annotations adopt a key-node evaluation scheme that assesses task completion at designated milestones rather than providing per-action ground-truth annotations with element-level targets.
Although the raw data provides per-action ground-truth annotations,~\footnote{\url{https://github.com/imeanai/webcanvas?tab=readme-ov-file\#download}} page sources are identified by URL without stored snapshots, and the referenced pages have since undergone content updates and layout modifications, making the original page structures unrecoverable.

\section{AXTree Preprocessing}
\label{apx:axtree_processing}

Our AXTree preprocessing follows BrowserGym~\citep{drouin2024workarena}, which extracts the accessibility tree via the Chrome DevTools Protocol, filters out nodes with no accessible content, and serializes each remaining node in an indentation-based text format (\texttt{[id] role name value}).~\footnote{\scriptsize \url{https://github.com/ServiceNow/BrowserGym/blob/main/browsergym/core/src/browsergym/core/observation.py}} We adopt this technique with three modifications.

First, each node is identified by a persistent identifier, the browser-assigned \texttt{backendDOMNodeId} or BrowserGym's \texttt{bid}, that remains stable across same-page DOM mutations.
This enables stable cross-step node matching and serves as the basis for the observation transition history in Section~\ref{sec:observation_change}.
Second, BrowserGym unconditionally removes all property-less \texttt{generic} and \texttt{none} nodes, which causes wrapper elements that group related content in the DOM to collapse into flat sibling lists.
We retain such a node when it has two or more child branches, each containing a visible descendant, preserving structural grouping that hierarchical decomposition relies on.
Finally, for \texttt{image} and \texttt{link} nodes whose accessible name is empty, the node is enriched with the corresponding \texttt{src} or \texttt{href} attribute retrieved from the DOM.

\section{Training Dataset Construction}
\label{apx:dataset_construction}

\subsection{Source Page Collection}

\paragraph{Domain categories.}
We select 10 domain categories from the 37 Tier 1 categories in the IAB Content Taxonomy 3.1 for their relevance to web agent tasks, covering Shopping, Travel, Technology \& Computing, Business and Finance, Education, Food \& Drink, Real Estate, Careers, Entertainment, and Sports.

\paragraph{Website selection.}
We use the Tranco top-1M ranking list snapshot from April 1, 2026 as the source.
To assign each website to a domain category, we embed the 37 Tier 1 
categories as reference embeddings and embed each website's concatenated title and description metadata using \texttt{sentence-transformers/paraphrase-MiniLM-L6-v2}.~\footnote{\url{https://huggingface.co/sentence-transformers/paraphrase-MiniLM-L6-v2}}
Each website is assigned to the nearest category by cosine similarity.
From the resulting clusters, we retain the 10 categories defined above and select the 500 highest-ranked websites by Tranco position across these categories.

\paragraph{Page URL sampling.}
For each website, page URLs are sampled from its \texttt{sitemap.xml} file.
Each URL is scored by the sum of three signals from the sitemap metadata, 
namely priority (0.0--1.0, default 0.5), change frequency (0.15 for daily 
or hourly, 0.1 for weekly), and URL depth (0.03 per path segment, up to 
5 levels).
Up to 100 URLs with the highest scores are retained per website.
Websites without an accessible \texttt{sitemap.xml} or unreachable via 
Playwright headless Chromium are excluded, removing 247 of the original 500.
The AXTree of each remaining page is extracted via Playwright, yielding 
21,974 pages from 253 websites.

\subsection{Data Annotation}

Since the knowledge of how web pages are functionally organized is implicit, we construct training data for both decomposition and abstraction using \texttt{gpt-5-mini-2025-08-27} as the annotator.
Because decomposition produces the region partition that abstraction then interprets, any partition error contaminates downstream abstraction labels.
We therefore add a verification stage between the two, retaining only pages where every region passes validation.
The annotation accordingly proceeds through three stages.

\paragraph{Decomposition annotation.}
The annotator receives each page's preprocessed AXTree together with the page URL and produces a list of region root node IDs.
Since the annotator occasionally assigns an entire page to a single region, a fallback mechanism re-partitions any region whose node count exceeds 50\% of the page total and is more than 10 times the median region size.
Of the 21,974 pages, 7 fail due to context length limits and 2,690 (12.2\%) trigger the fallback.
The remaining 21,967 pages yield 547,075 regions, averaging 24.9 per page.

\paragraph{Partition verification.}
The annotator receives each page's region partition and identifies regions that were incorrectly decomposed.
Since even a single invalid region is sufficient to corrupt the abstraction 
labels derived from it, we retain only pages that yield a valid region partition, one in which every region is correctly decomposed.
This reduces 21,967 pages to 2,052 (9.3\%) with 46,487 regions, averaging 22.7 per page.

\paragraph{Abstraction annotation.}
The annotator receives each verified region's AXTree subtree and produces a purpose and a state summary.
Of the 46,487 regions, 1,340 consist solely of \texttt{none} or \texttt{generic} nodes with no visible content and are excluded, yielding 45,147 annotated regions from 2,052 pages.
The annotations reduce the average region representation from 176.6 tokens to 56.2 tokens under the OpenAI \texttt{o200k\_base} tokenizer, resulting in a 68.2\% reduction.

Table~\ref{tab:dataset_stats} summarizes the dataset at each stage of the construction process. The annotation prompts used at each stage are provided in Figure~\ref{pmt:decomposition}, \ref{pmt:verification}, and \ref{pmt:abstraction}, respectively.

\begin{table}[htbp]
\centering
\caption{Statistics at each stage of training dataset construction.}
\label{tab:dataset_stats}
\footnotesize
\begin{tabular}{lrrrrrr}
\toprule
Domain category & \# Sites & \# Source pages & \# Verified pages & \# Edges & \# Cut edges & \# Regions \\
\midrule
Technology \& Computing & 97 & 8,329 & 674 (32.85\%) & 234,708 & 15,366 (6.55\%) & 15,677 \\
Entertainment & 34 & 3,103 & 254 (12.38\%) & 141,090 & 9,363 (6.64\%) & 9,428 \\
Sports & 34 & 3,028 & 291 (14.18\%) & 91,487 & 6,676 (7.30\%) & 6,861 \\
Business and Finance & 25 & 2,293 & 274 (13.35\%) & 50,644 & 4,836 (9.55\%) & 4,968 \\
Shopping & 19 & 1,632 & 266 (12.96\%) & 17,471 & 1,823 (10.43\%) & 1,986 \\
Education & 19 & 1,523 & 102 (4.97\%) & 26,764 & 2,488 (9.30\%) & 2,535 \\
Real Estate & 10 & 807 & 63 (3.07\%) & 17,581 & 1,158 (6.59\%) & 1,209 \\
Travel & 7 & 509 & 29 (1.41\%) & 5,208 & 574 (11.02\%) & 565 \\
Careers & 6 & 600 & 44 (2.14\%) & 24,655 & 1,503 (6.10\%) & 1,535 \\
Food \& Drink & 2 & 150 & 55 (2.68\%) & 7,346 & 329 (4.48\%) & 383 \\
\midrule
\textbf{Total} & 253 & 21,974 & \textbf{2,052} (100\%) & \textbf{616,954} & \textbf{44,116} (7.15\%) & \textbf{45,147} \\
\bottomrule
\end{tabular}
\end{table}

\section{\framework{} Implementation Details}
\label{apx:framework_implementation}

\subsection{Hierarchical Decomposition}

\paragraph{Node features.}
The feature vector $\mathbf{x}_v$ is 16-dimensional, concatenating a learned role embedding (11 dimensions) with five numeric features.
The role vocabulary contains 204 entries, 203 from the Chromium accessibility role enumeration~\footnote{Chromium 125.0.6422.26. \texttt{chromium/src/ui/accessibility/ax\_enums.mojom}} and one for unknown roles.
The five numeric features are the node's depth in the tree, subtree size, number of children, accessible name presence, and child role diversity, providing structural cues beyond the role embedding for boundary classification.
Accessible name presence is set to 1 when the node has a non-empty accessible name and 0 otherwise.
Child role diversity calculates the ratio of unique child roles to the number of children.

\paragraph{Model architecture.}
\textsc{RegionEncoder} and \textsc{EdgeClassifier} are both three-layer MLPs with ReLU activations and a hidden dimension of 256.
\textsc{RegionEncoder} maps a 272-dimensional input ($\mathbf{x}_v$ and the merged children aggregation) to the 256-dimensional representation $\mathbf{r}_v$.
\textsc{EdgeClassifier} maps a 528-dimensional input ($\mathbf{x}_v$, $\mathbf{r}_{c_i}$, and $\bar{\mathbf{r}}$) to a scalar logit $\hat{y}_{v,c_i}$.
The model totals approximately 536K parameters including the role embedding table.
The full inference procedure is given in Algorithm~\ref{alg:decomposition}.

\begin{algorithm}[htbp]
\caption{Hierarchical Decomposition}
\label{alg:decomposition}
\begin{algorithmic}[1]
\Require Page AXTree $\mathcal{T} = (\mathcal{V}, \mathcal{E})$, threshold $\tau$
\Ensure Region partition $\mathcal{R}$
\State $\mathcal{R} \gets \emptyset$
\For{each node $v \in \mathcal{V}$ in bottom-up order}
\State $S_v \gets \{v\}$
\State $\mathbf{x}_v \gets \text{Concat}(\mathbf{E}_{\text{role}}(v),\; \mathbf{n}_v)$
\If{$v$ is a leaf}
\State $\mathbf{r}_v \gets \textsc{RegionEncoder}(\mathbf{x}_v,\; \mathbf{0})$
\Else
\State Let $c_1, \ldots, c_k$ be the children of $v$
\State $\bar{\mathbf{r}}_v \gets \frac{1}{k}\sum_{j=1}^{k} \mathbf{r}_{c_j}$
\Comment{Sibling mean}
\State $\mathcal{M}_v \gets \emptyset$
\For{$i = 1, \ldots, k$}
\If{$\textsc{EdgeClassifier}(\mathbf{x}_v,\; \mathbf{r}_{c_i},\; \bar{\mathbf{r}}_v) \geq \tau$}
\State $\mathcal{R} \gets \mathcal{R} \cup \{S_{c_i}\}$
\Comment{Cut: $c_i$'s subtree constitutes a region}
\Else
\State $\mathcal{M}_v \gets \mathcal{M}_v \cup \{c_i\}$
\State $S_v \gets S_v \cup S_{c_i}$
\Comment{Merge: $c_i$'s subtree merges into $v$'s region}
\EndIf
\EndFor
\If{$\mathcal{M}_v \neq \emptyset$}
\State $\mathbf{r}_v \gets \textsc{RegionEncoder}\bigl(\mathbf{x}_v,\; \tfrac{1}{|\mathcal{M}_v|}\sum_{c_j \in \mathcal{M}_v} \mathbf{r}_{c_j}\bigr)$
\Else
\State $\mathbf{r}_v \gets \textsc{RegionEncoder}(\mathbf{x}_v,\; \mathbf{0})$
\EndIf
\EndIf
\EndFor
\State $\mathcal{R} \gets \mathcal{R} \cup \{S_{v_{\text{root}}}\}$
\Comment{Root subtree constitutes the final region}
\State \Return $\mathcal{R}$
\end{algorithmic}
\end{algorithm}

\paragraph{Training configuration.}
The model is trained with teacher forcing, where ground-truth edge labels determine cut and merge decisions during the bottom-up traversal rather than the model's own predictions.
Training runs for 140 epochs on a NVIDIA RTX A6000 GPU with Adam optimizer at a learning rate of $1 \times 10^{-4}$ and gradient clipping at $1.0$, and focal 
loss with $\alpha = 0.75$ and $\gamma = 2.0$ to address the class 
imbalance between merge and cut edges.
The data is split into 90\% training and 10\% validation sets at the page level with seed 42.

\paragraph{Checkpoint selection and threshold tuning.}
The training epoch and the inference threshold $\tau$ are determined in two steps, each using the metric that matches its objective.

\begin{figure}[htbp]
\centering
\begin{minipage}[c]{0.45\linewidth}
\centering
\includegraphics[width=\textwidth]{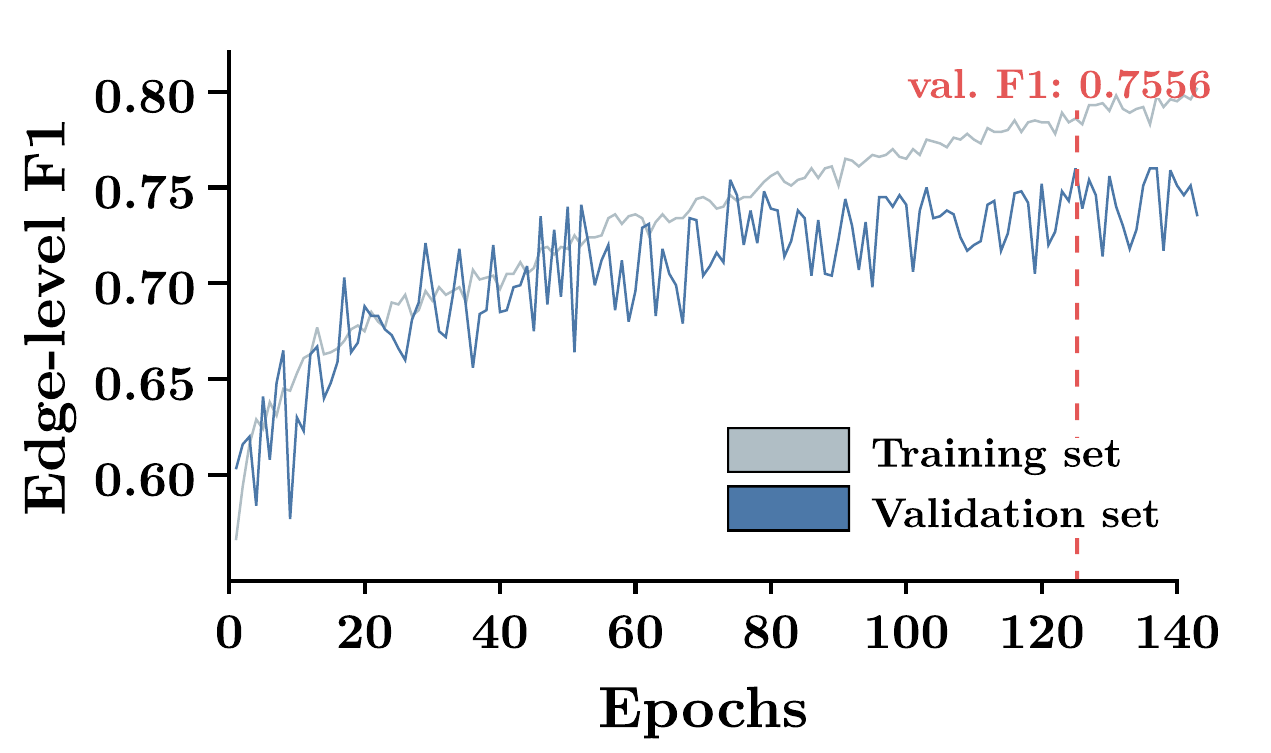}
\captionof{figure}{Edge-level F1 on training and validation sets over 140 epochs. Epoch 125 is selected for deployment.}
\label{fig:decomposition_epochs}
\end{minipage}
\hfill
\begin{minipage}[c]{0.5\linewidth}
\centering
\captionof{table}{Region-level precision, recall, and F1 across inference thresholds at epoch 125. $\tau = 0.55$ achieves the highest F1.}
\label{tab:decomposition_threshold}
\small
\begin{tabular}{cccc}
\toprule
$\tau$ & Precision & Recall & F1 \\
\midrule
0.35 & 0.5035 & 0.8440 & 0.6307 \\
0.40 & 0.5704 & 0.8438 & 0.6806 \\
0.45 & 0.6331 & 0.8315 & 0.7189 \\
0.50 & 0.7185 & 0.8082 & 0.7607 \\
\textbf{0.55} & \textbf{0.7755} & \textbf{0.7743} & \textbf{0.7749} \\
0.60 & 0.7964 & 0.6852 & 0.7366 \\
0.65 & 0.8341 & 0.6139 & 0.7073 \\
0.70 & 0.8589 & 0.5396 & 0.6628 \\
\bottomrule
\end{tabular}
\end{minipage}
\end{figure}

The training epoch is selected based on edge-level F1 on the validation set, as the model directly optimizes edge-level binary classification during training.
Among the epochs with the highest validation F1, we choose the one with the smallest training-validation F1 gap to avoid overfitting, yielding epoch 125.
Figure~\ref{fig:decomposition_epochs} shows the edge-level F1 curves over training.

The inference threshold $\tau$ converts edge-level logits into a region partition, whose quality is not captured by edge-level metrics.
We therefore tune $\tau$ at the region level.
Each ground-truth region is matched to the predicted region with the highest Intersection-over-Union (IoU), and counted as matched if this IoU meets or exceeds 0.5.
Region-level precision, recall, and F1 are then computed over the matched counts relative to the total predicted and ground-truth regions.
This yields $\tau = 0.55$.
Table~\ref{tab:decomposition_threshold} reports the region-level metrics across threshold values.

\subsection{Semantic Abstraction}

\paragraph{Training configuration.}
\texttt{Qwen3-0.6B} is fine-tuned with full supervised fine-tuning in \texttt{bfloat16} precision with gradient checkpointing.
Each training example pairs a region's preprocessed AXTree subtree as input with a JSON object containing the corresponding purpose $p_i$ and state summary $s_i$ as output, using the annotation prompt as the instruction prefix, which is shown in Figure~\ref{pmt:abstraction}.
The loss is computed only on the output tokens, with all input and padding tokens masked.
Training runs for 90 epochs (76,200 steps) on 3 NVIDIA RTX A6000 GPUs using distributed data parallel (DDP)~\citep{li2020ddp} with a per-device batch size of 1 and gradient accumulation of 16, yielding an effective batch size of 48.
The optimizer is AdamW with a learning rate of $5 \times 10^{-6}$, $200$ linear warmup steps, and cosine decay.
The maximum sequence length is 8,192 tokens, and 37 samples (0.08\%) exceeding this limit are skipped during training.
The data is split into 90\% training and 10\% validation sets with the same seed 42 used throughout decomposition model training.

\begin{figure}[htbp]
\centering
\includegraphics[width=0.63\linewidth]{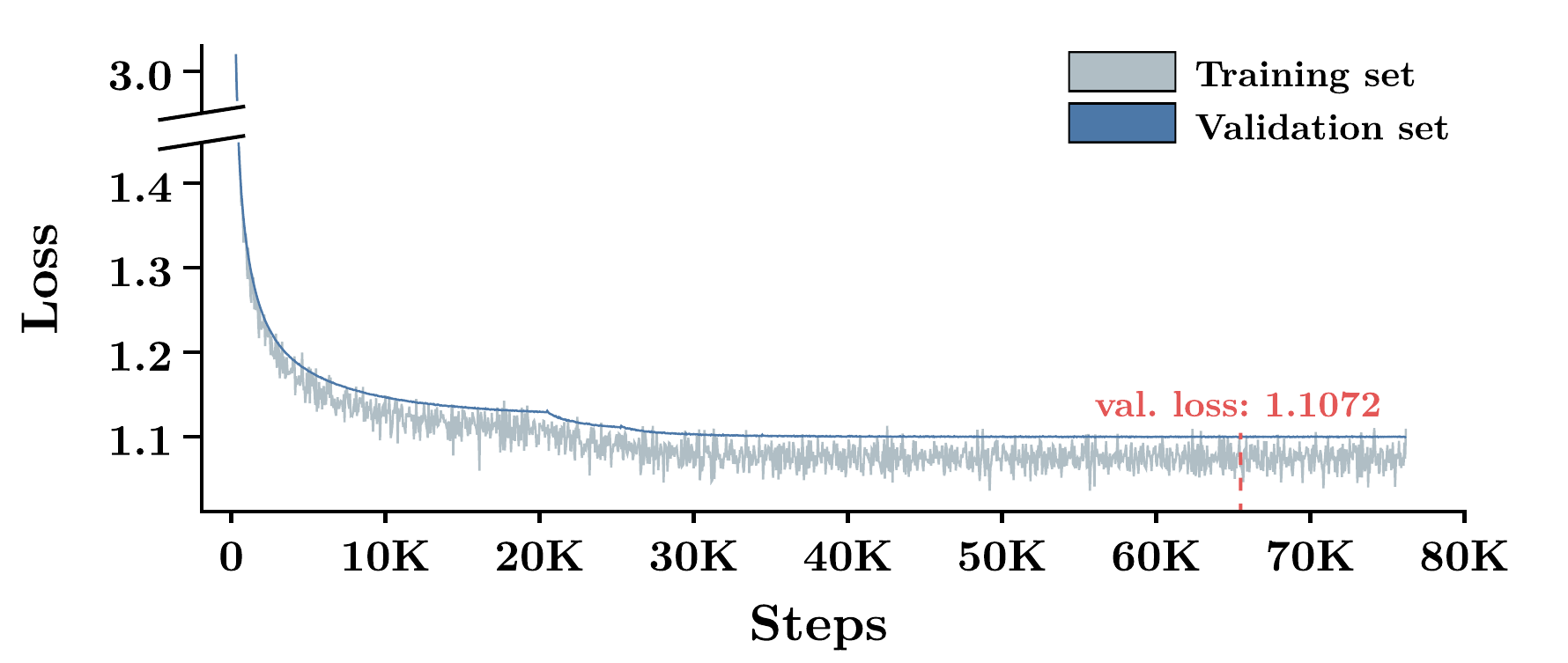}
\caption{Training and validation loss over 90 epochs. Step 65,350 is selected for deployment.}
\label{fig:abstraction_steps}
\end{figure}

\paragraph{Checkpoint selection.}
The checkpoint at step 65,350 is selected by jointly considering validation loss and manual quality assessment of sampled outputs.
Figure~\ref{fig:abstraction_steps} shows the training and validation loss curves.

\paragraph{Inference.}
The fine-tuned model processes regions with greedy decoding.
The annotation prompt as in Figure~\ref{pmt:abstraction} is reused at inference to maintain distributional consistency between training and deployment.

\section{Case Studies}
\label{apx:case_study}
We provide qualitative case studies of \framework{}'s decomposition and abstraction stage on representative pages from each WebArena domain in Tables~\ref{tab:case_shopping} through~\ref{tab:case_map}.

\section{Prompts}
\label{apx:prompts}

For reproducibility, we provide all prompts used in this work in Figures~\ref{pmt:decomposition} through~\ref{pmt:action_pipeline}.

\begin{table}[ht]
\caption{\framework{} output on WebArena Shopping domain.}
\label{tab:case_shopping}
\begin{tabular*}{\linewidth}{@{}p{\linewidth}@{}}
\toprule
\textbf{Page URL}: http://localhost:7770 (343 nodes)\\
\midrule
\textbf{Hierarchical Decomposition}: total 23 regions\\
\midrule
\multicolumn{1}{c}{\includegraphics[width=0.97\linewidth]{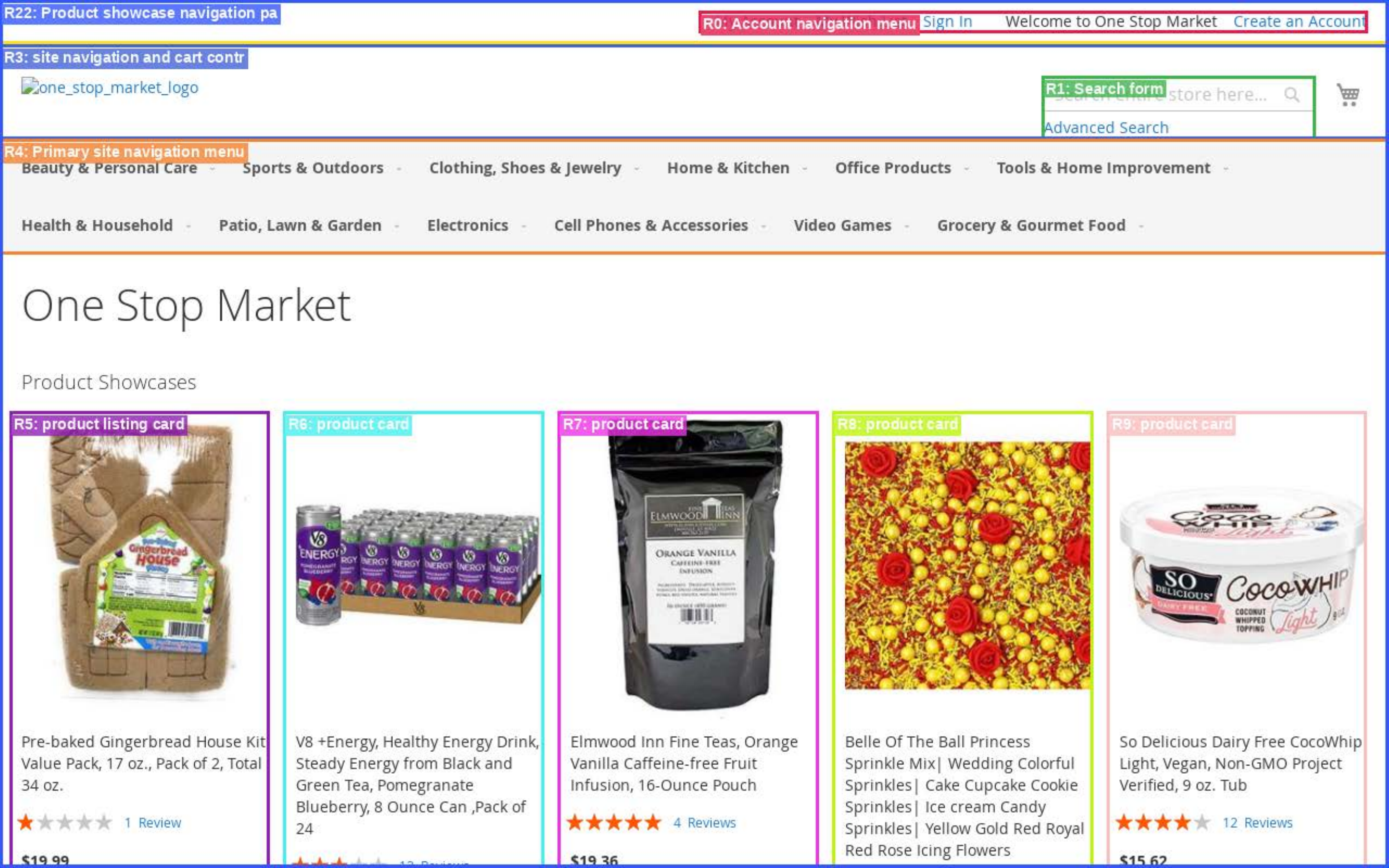}}\\
\midrule
\textbf{Semantic Abstraction}: R0, R1, R5, R7, R8\\
\midrule
\textbf{R0} (\textbf{purpose}: Account navigation menu)\\
\textbf{state summary}: Provides navigation links to account-related pages (My Account, My Wish List, Sign In, Create an Account) and a Welcome message. The Create an Account link is actionable to initiate a new account.\\
\midrule
\textbf{R1} (\textbf{purpose}: Search form)\\
\textbf{state summary}: Search is currently enabled with a combobox labeled "Search" and a "Advanced Search" link. The combobox is not expanded and the button is disabled.\\
\midrule
\textbf{R5} (\textbf{purpose}: product listing card)\\
\textbf{state summary}: Product: Pre-baked Gingerbread House Kit Value Pack, 17 oz., Pack of 2, Total 34 oz. with a 20\% rating and \$19.99. Available actions: Add to Cart, Add to Wish List, and Add to Compare.\\
\midrule
\textbf{R6} (\textbf{purpose}: product card)\\
\textbf{state summary}: Healthy energy drink with a 57\% rating and \$14.47 price. Available actions: Add to Cart, Add to Wish List, and Add to Compare.\\
\midrule
\textbf{R7} (\textbf{purpose}: product card)\\
\textbf{state summary}: Product: Elmwood Inn Fine Teas, Orange Vanilla Caffeine-free Fruit Infusion, 16-Ounce Pouch (95\% rating) priced at \$19.36. Available actions: Add to Cart, Add to Wish List, and Add to Compare.\\
\bottomrule
\end{tabular*}
\end{table}

\begin{table}[ht]
\caption{\framework{} output on WebArena CMS (shopping admin) domain.}
\label{tab:case_cms}
\begin{tabular*}{\linewidth}{@{}p{\linewidth}@{}}
\toprule
\textbf{Page URL}: http://localhost:7780/admin/admin/dashboard (217 nodes)\\
\midrule
\textbf{Hierarchical Decomposition}: total 17 regions\\
\midrule
\multicolumn{1}{c}{\includegraphics[width=0.97\linewidth]{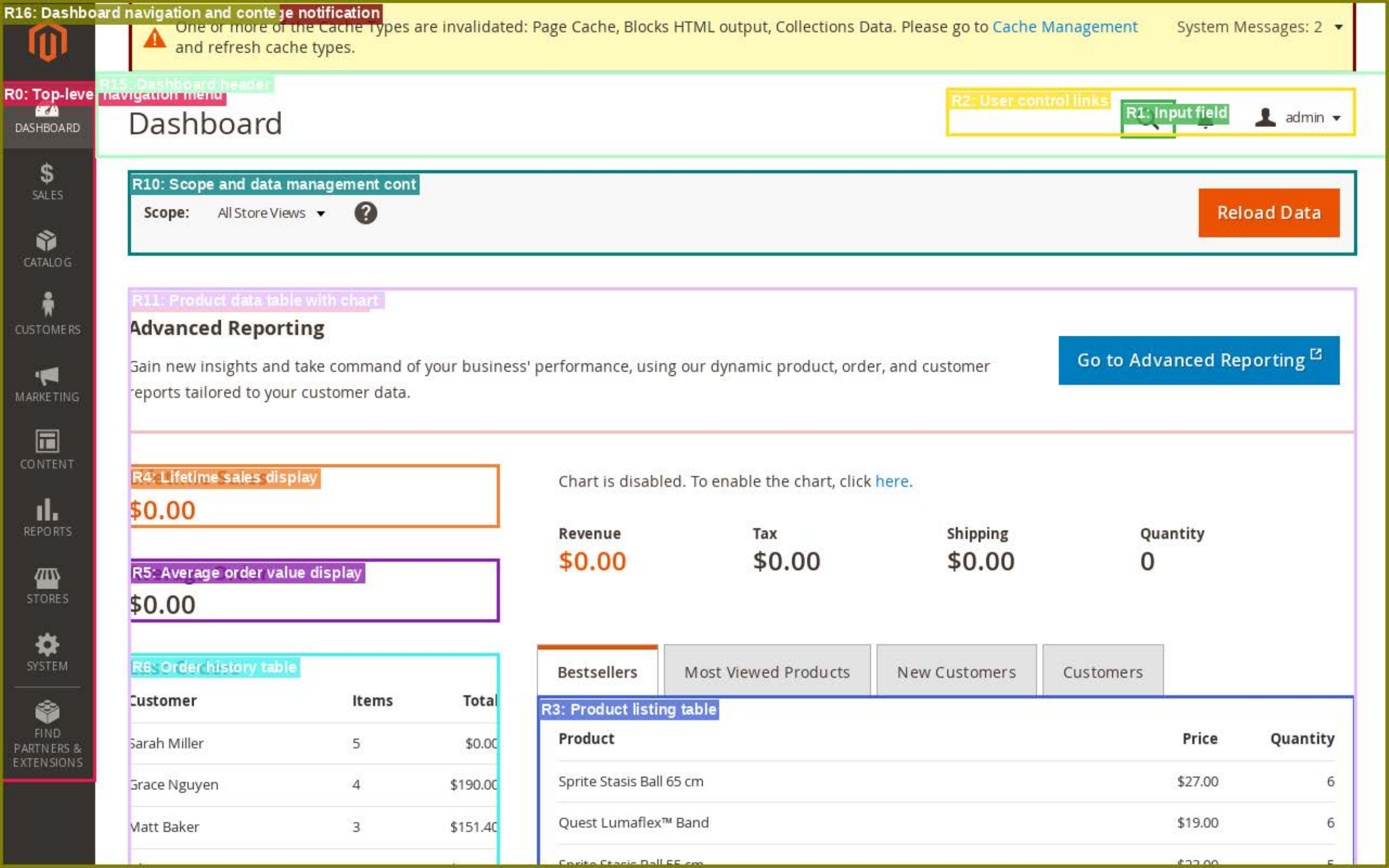}}\\
\midrule
\textbf{Semantic Abstraction}: R2, R5, R6, R10\\
\midrule
\textbf{R2} (\textbf{purpose}: User control links)\\
\textbf{state summary}: Contains two clickable links labeled "admin"s. Clicking either link navigates to the corresponding admin page.\\
\midrule
\textbf{R5} (\textbf{purpose}: Average order value display)\\
\textbf{state summary}: Shows an average order value of \$0.00. No interactive controls are present in this region.\\
\midrule
\textbf{R6} (\textbf{purpose}: Order history table)\\
\textbf{state summary}: Shows order details for five orders (ID 299, 65, 125, 136, 230) with each row showing customer name, item count, and total. Each order link is actionable (clickable URL) to view the order.\\
\midrule
\textbf{R10} (\textbf{purpose}: Scope and data management controls)\\
\textbf{state summary}: Shows a 'Scope:' heading and provides a 'All Store Views' button with a menu popup and a 'Reload Data' button. The 'What is this?' link is actionable for clarification.\\
\bottomrule
\end{tabular*}
\end{table}

\begin{table}[ht]
\caption{\framework{} output on WebArena Reddit domain.}
\label{tab:case_reddit}
\begin{tabular*}{\linewidth}{@{}p{\linewidth}@{}}
\toprule
\textbf{Page URL}: http://localhost:9999/friedly-reminder-bookshop-org-exists (4,151 nodes)\\
\midrule
\textbf{Hierarchical Decomposition}: total 345 regions\\
\midrule
\multicolumn{1}{c}{\includegraphics[width=0.97\linewidth]{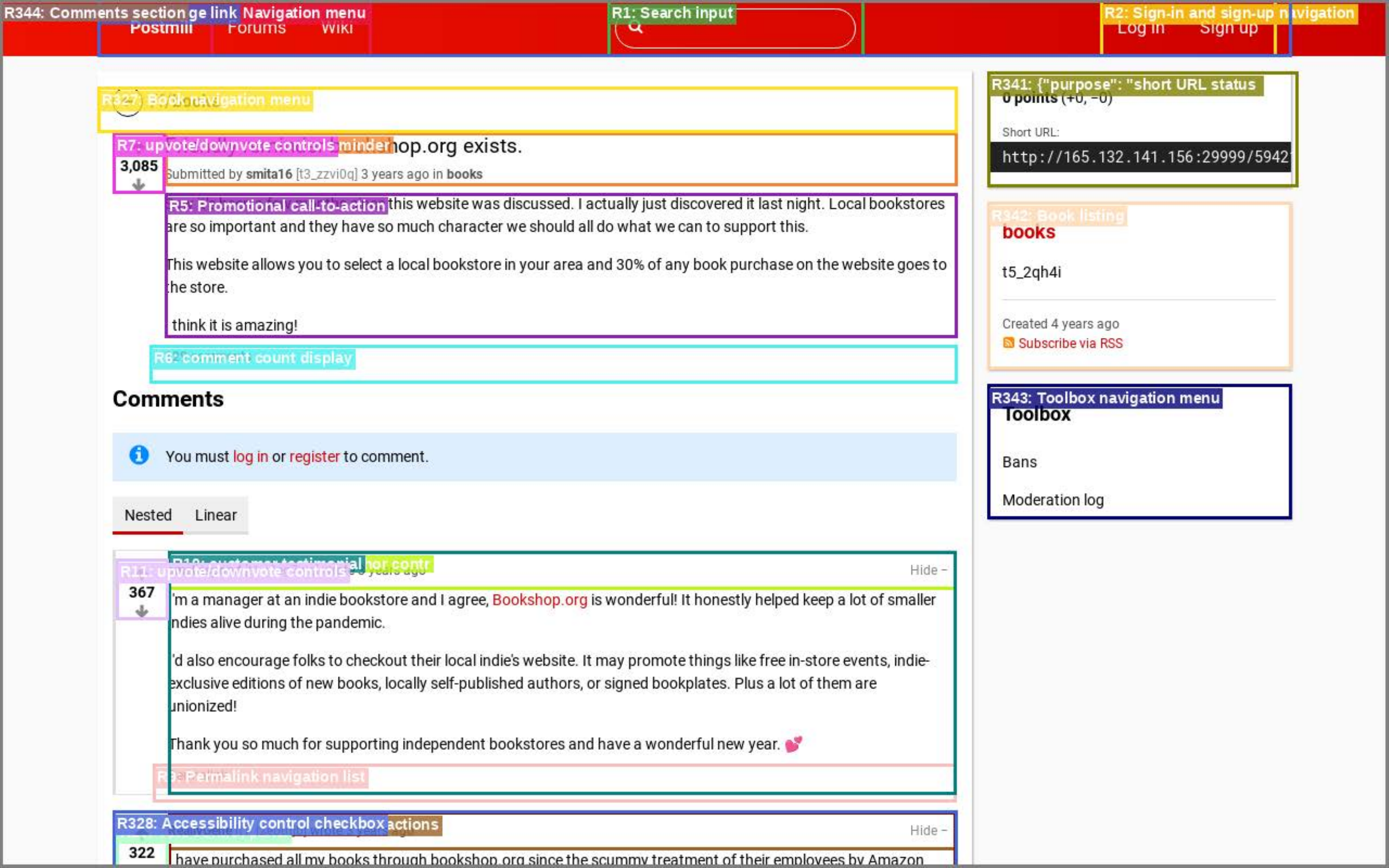}}\\
\midrule
\textbf{Semantic Abstraction}: R5, R6, R11, R342\\
\midrule
\textbf{R5} (\textbf{purpose}: Promotional call-to-action)\\
\textbf{state summary}: Promotes a local bookstore program that 30\% of book purchases go to the store and encourages supporting local bookstores. The region contains static text and a closing statement that appears to be a call to action.\\
\midrule
\textbf{R6} (\textbf{purpose}: comment count display)\\
\textbf{state summary}: Shows a count of 129 comments. The item is a link that can be activated to open the comment list or view more details.\\
\midrule
\textbf{R11} (\textbf{purpose}: upvote/downvote controls)\\
\textbf{state summary}: Contains two buttons labeled "Upvote" and "Downvote" with a numeric value of 367 displayed. Clicking the buttons will toggle the up/down vote state and the 367 number is static text showing the current count.\\
\midrule
\textbf{R342} (\textbf{purpose}: Book listing)\\
\textbf{state summary}: Contains a single book entry with a 'books' link and a 'Subscribe via RSS' image. The book's timestamp shows it was created 4 years ago.\\
\bottomrule
\end{tabular*}
\end{table}

\begin{table}[ht]
\caption{\framework{} output on WebArena Gitlab domain.}
\label{tab:case_gitlab}
\begin{tabular*}{\linewidth}{@{}p{\linewidth}@{}}
\toprule
\textbf{Page URL}: http://localhost:8023/byteblaze/a11y-syntax-highlighting (546 nodes)\\
\midrule
\textbf{Hierarchical Decomposition}: total 14 regions\\
\midrule
\multicolumn{1}{c}{\includegraphics[width=0.97\linewidth]{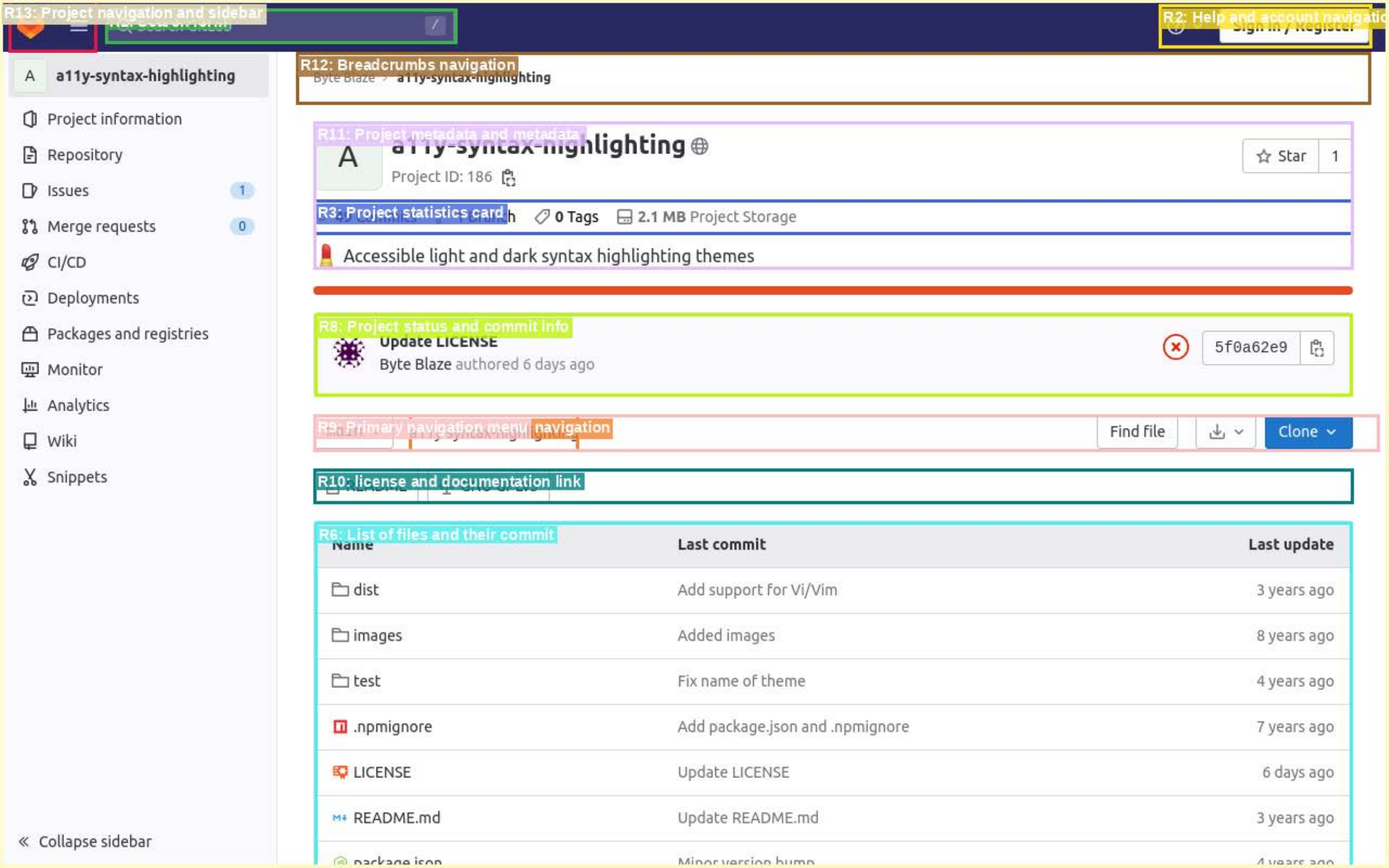}}\\
\midrule
\textbf{Semantic Abstraction}: R2, R3, R6, R10\\
\midrule
\textbf{R2} (\textbf{purpose}: Help and account navigation links)\\
\textbf{state summary}: Contains two links: a 'Help' link with an image and a 'Sign in / Register' link. Clicking either will navigate to the help documentation or account sign-in/register page.\\
\midrule
\textbf{R3} (\textbf{purpose}: Project statistics card)\\
\textbf{state summary}: Shows project statistics: 49 commits, 1 branch, 0 tags, and 2.1 MB project storage. All items are clickable links that navigate to the corresponding metrics.\\
\midrule
\textbf{R6} (\textbf{purpose}: List of files and their commit/updates)\\
\textbf{state summary}: Contains a list of files (dist, images, test, LICENSE, README.md, package.json) with their last commit and update dates. Each file is a link that opens the file's page or shows the file's name and time.\\
\midrule
\textbf{R10} (\textbf{purpose}: license and documentation links)\\
\textbf{state summary}: Contains two clickable links: a README image and a GNU GPLv3 license link. Click either link to open the corresponding documentation or license page.\\
\bottomrule
\end{tabular*}
\end{table}

\begin{table}[ht]
\caption{\framework{} output on WebArena Map domain.}
\label{tab:case_map}
\begin{tabular*}{\linewidth}{@{}p{\linewidth}@{}}
\toprule
\textbf{Page URL}: https://www.openstreetmap.org/directions?engine=fossgis\_osrm\_car... (176 nodes)\\
\midrule
\textbf{Hierarchical Decomposition}: total 13 regions\\
\midrule
\multicolumn{1}{c}{\includegraphics[width=0.97\linewidth]{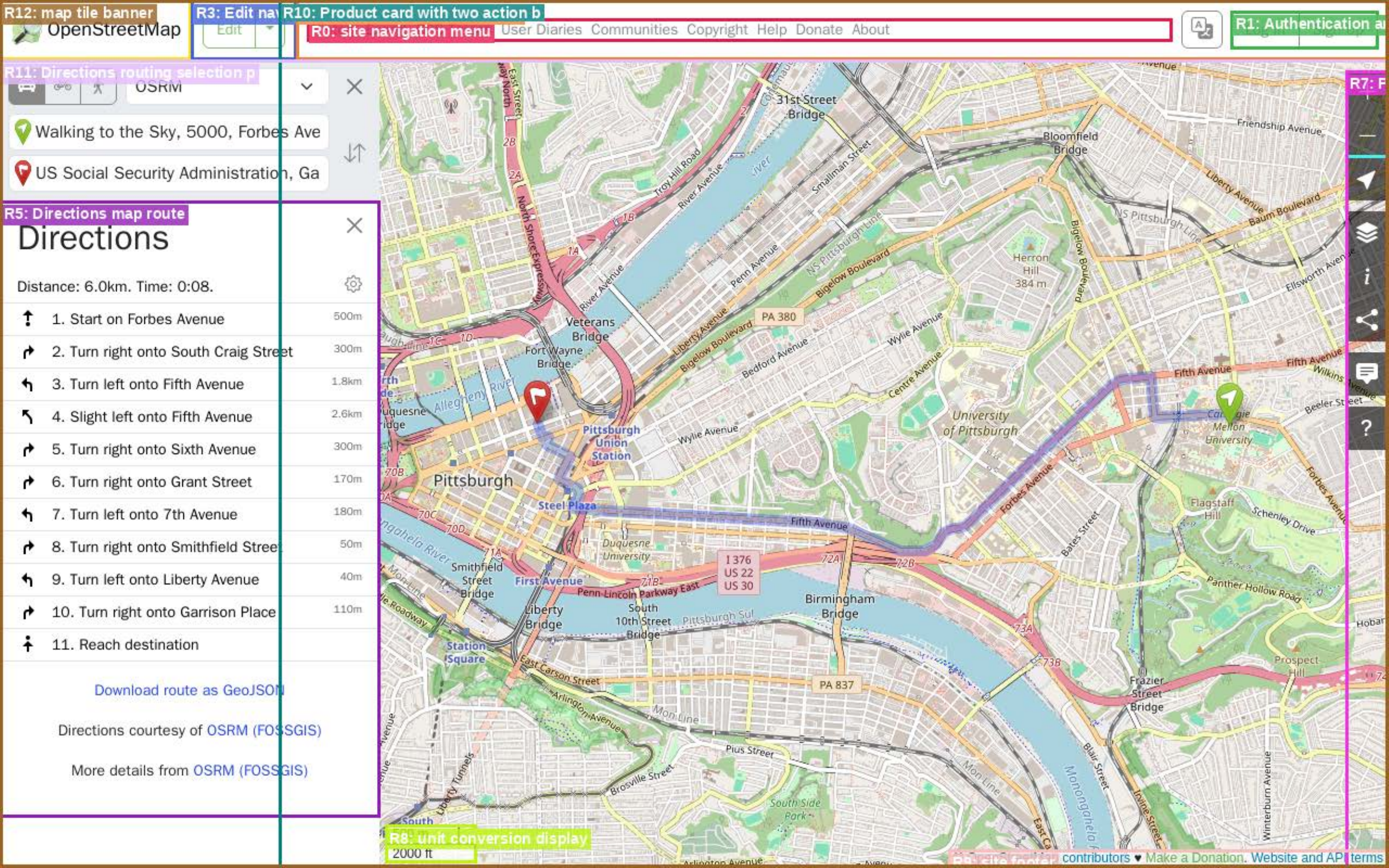}}\\
\midrule
\textbf{Semantic Abstraction}: R0, R1, R5, R7, R11\\
\midrule
\textbf{R0} (\textbf{purpose}: site navigation menu)\\
\textbf{state summary}: Provides navigational links to site sections: History, Export, GPS Traces, User Diaries, Communities, Copyright, Help, Donate, and About. Each item is a clickable link to navigate to the corresponding page.\\
\midrule
\textbf{R1} (\textbf{purpose}: Authentication and sign-up navigation)\\
\textbf{state summary}: Provides two navigation links: 'Log In' to initiate account access and 'Sign Up' to create a new account. Both links are actionable and can be activated to proceed with the respective authentication or sign-up process.\\
\midrule
\textbf{R5} (\textbf{purpose}: Directions map route)\\
\textbf{state summary}: Provides a route map with 11 steps (1–11) and a destination. Includes a downloadable GeoJSON file and a link to the OSRM (FOSSGIS) source. The heading 'Directions' is a heading and the table shows distance and time for each step.\\
\midrule
\textbf{R7} (\textbf{purpose}: Page header controls)\\
\textbf{state summary}: Provides navigation links to Layers, Legend, Share, Add a note to the map, and Query features. A 'Show My Location' button is available to open the location view.\\
\midrule
\textbf{R11} (\textbf{purpose}: Directions routing selection panel)\\
\textbf{state summary}: Selects directions services (OSRM) and provides a 'Reverse Directions' button to reverse the route. The 'From' and 'To' fields are populated with the specified addresses and the 'Close' button cancels the panel.\\
\bottomrule
\end{tabular*}
\end{table}

\begin{figure}[ht]
\begin{tcolorbox}[colback=white, colframe=blue!70, title={Decomposition prompt}]
\footnotesize
You are an observation space analyst for web agents. You partition a web page's accessibility tree (AXTree) into non-overlapping functional regions so that an autonomous agent can understand and interact with the page.\\

<definition>\\
A functional region is a subtree whose elements are collectively organized to serve a distinct purpose. Functional purpose is not a property of individual elements. It arises from how elements are collectively organized. A single link has an element-level action, but a region-level purpose emerges only when multiple elements are organized together to fulfill a coherent function. Every node belongs to exactly one region.\\
</definition>\\

<constraints>\\
1. Structural containers (the tree root, ARIA landmarks like banner/main/contentinfo) group content by page position, not by purpose. Always evaluate their children. A container becomes a region only for children that do not form their own regions.\\

2. A region must be meaningful to an agent, something it would need to independently recognize or interact with to carry out a task. Purely decorative elements and isolated utility shortcuts belong to their parent's region.\\
</constraints>\\

<algorithm>\\
To evaluate a node N, apply these steps in order.\\

Step 1. Container passthrough.\\
If N is the tree root or an ARIA landmark, evaluate each direct child of N by applying this algorithm recursively. N itself becomes a region only for its remaining children, those that did not form their own regions. If all children form regions, N is not recorded.\\

Step 2. Evaluate N.\\
For each candidate N that is not a container:\\

(a) If N's children are structurally repetitive (a collection of entries), determine whether each entry's purpose arises from its own internal organization: its own elements collectively organized into different roles (information, metadata, and actions) around a single entity, interpretable without shared context from siblings or parent structure. If yes, each entry is its own region. Apply this algorithm recursively to each. If no, N is one region. Record N.\\

(b) If N has multiple children that serve distinct purposes and each child's purpose arises from its own internal organization, each such child is its own region. Apply this algorithm recursively to each. Children whose purpose does not arise from their own internal organization remain in N's region.\\

(c) If neither (a) nor (b) applies, N serves one coherent purpose. Record N.\\

Step 3. Meaningfulness check.\\
Before recording N, verify it is meaningful to an agent (see constraint 2). If not, N belongs to its parent's region.\\
</algorithm>\\

<output\_format>\\
Output the recorded region root node IDs as a comma-separated list.\\
After the list, output nothing further.\\
</output\_format>\\

URL: \{url\}\\

AXTree:\\
\{axtree\}
\end{tcolorbox}
\caption{Prompt for decomposition annotation stage.}
\label{pmt:decomposition}
\end{figure}

\begin{figure}[ht]
\begin{tcolorbox}[colback=white, colframe=blue!70, title={Verification prompt}]
\footnotesize
You are an observation space analyst for web agents. You verify whether a page's accessibility tree (AXTree) has been correctly partitioned into functional regions.\\

<definition>\\
A functional region is a subtree whose elements are collectively organized to serve a distinct purpose. Functional purpose is not a property of individual elements. It arises from how elements are collectively organized. A single link has an element-level action, but a region-level purpose emerges only when multiple elements are organized together to fulfill a coherent function.\\
</definition>\\

<criteria>\\
For each region, verify that it is correctly formed by checking:\\

1. The region corresponds to one recognizable functional unit on the page, an area that an agent would identify as serving a single role.\\

2. If you can identify multiple sub-components within the region that are each independently recognizable as their own functional area on the page, the region is incorrectly formed. Those areas should be separate regions.\\
</criteria>\\

<output\_format>\\
Output the IDs of incorrectly formed regions as a comma-separated list (e.g., R3, R7).\\
If no region is incorrectly formed, output: none\\
After the output, output nothing further.\\
</output\_format>\\

URL: \{url\}\\

\{region\_partition\}
\end{tcolorbox}
\caption{Prompt for partition verification stage.}
\label{pmt:verification}
\end{figure}

\begin{figure}[ht]
\begin{tcolorbox}[colback=white, colframe=blue!70, title={Abstraction prompt}]
\footnotesize
You are an observation space analyst for web agents. You produce semantic descriptions of functional regions extracted from web page accessibility trees (AXTree).\\

<task>\\
Given a region's AXTree subtree, produce two descriptions:\\

1. purpose: Identify the collective function that the region's elements are organized to serve. This should name the type of region, not describe its current contents or enumerate its features. Write a short noun phrase.\\

2. state\_summary: Interpret the region's current content and available actions to inform task-based decision making. Lead with the key information an agent would match against a task, not with descriptions of what the region shows. Write one to two concise sentences.\\
</task>\\

<guidelines>\\
- Derive both fields solely from the elements present in the subtree.\\
- For purpose, identify what the elements collectively accomplish, not what individual elements are.\\
- For state\_summary, interpret and select what matters for decision making, not exhaustively describe elements.\\
- Always output in English, translating non-English content as needed.\\
</guidelines>\\

\{region\_axtree\}\\
\end{tcolorbox}
\caption{Prompt for abstraction, used across annotation, training, and inference.}
\label{pmt:abstraction}
\end{figure}

\begin{figure}[ht]
\begin{tcolorbox}[colback=white, colframe=blue!70, title={Selection prompt}]
\footnotesize
You are a page region selector for a web agent. You receive the agent's task, the actions taken so far, and a list of functional regions on the current page, each described by its purpose and state summary. Select the regions the agent needs on this page to make progress on the task.\\

<principles>\\
1. First understand what the current page offers from the full set  of region abstractions. Then, given the task and the action history, select every region whose content could be relevant to the task, whether the agent needs to interact with it or read information from it. Do not exclude regions based on an assumed course of action.\\

2. Exclude a region only when its purpose is clearly unrelated to the task. If relevance cannot be determined from the description, include the region.\\

3. When multiple regions share a similar purpose and their state summaries do not indicate which ones the task requires, include all of them.\\

4. A state summary that appears to match the task does not by itself justify excluding other potentially relevant regions. The rendered content may not satisfy the task's exact requirements.\\
</principles>\\

<output\_format>\\
Output the selected region IDs as a comma-separated list (e.g., R3, R7).\\
After the list, output nothing further.\\
</output\_format>\\

Task: \{task\_instruction\}\\

Action history:\\
\{action\_history\}\\

\{region\_abstractions\}
\end{tcolorbox}
\caption{Prompt for task-relevant region selection.}
\label{pmt:selection}
\end{figure}

\begin{figure}[ht]
\begin{tcolorbox}[colback=white, colframe=blue!70, title={Action selection prompt}]
\footnotesize
You are a web agent. You receive a task, the current page state, and your previous actions. You select the next action to make progress on the task.\\

<action\_space>\\
\{action\_space\}\\
<action\_space>\\

Elements are identified by unique bid. Use bid to refer to elements in your actions. Interacting with comboboxes, dropdowns, and auto-complete fields may require different actions depending on the element. Try select\_option first. If it does not work, try fill or click and observe the result. Your final message to the user must contain only the answer value in the format implied by the task, with no prefixes, explanations, or restatements.\\
</action\_space>\\

<output\_format>\\
Reason about the current state and decide the next action inside <think> tags.\\
Then output exactly one action inside <action> tags.\\
After the tags, output nothing further.\\

<think>\\
Your step-by-step reasoning.\\
</think>\\
<action>\\
click('a324')\\
</action>\\
</output\_format>\\

Task: \{task\_instruction\}\\

Action history:\\
\{action\_history\}\\

\{axtree\}
\end{tcolorbox}
\caption{Prompt template for action selection.
\{action\_space\} is replaced with the set of 15 available actions and their descriptions provided by BrowserGym. (e.g., \texttt{click}, \texttt{fill})}
\label{pmt:action}
\end{figure}

\begin{figure}[ht]
\begin{tcolorbox}[colback=white, colframe=blue!70, title={Action selection prompt (w/ \pipeline)}]
\footnotesize
\# action space\\
view\_all()\\
\hspace*{1em}Description: Reorganize the current page state into functional regions and reveal all of them in their full AXTree form. Use this when the currently exposed regions appear insufficient for the task, for example when the target element seems missing from the observation or when no available action makes progress. This action remains in effect until the agent navigates to a new page.\\
\hspace*{1em}Examples:\\
\hspace*{2em}view\_all()\\

\# observation space\\
<observation\_space>\\
The page is split into functional regions, each rendered as a <Rx purpose="..."> ... </Rx> block. Regions deemed task-relevant render their full AXTree subtree inside the block; others appear as the opening tag with no inner content. Newly appeared nodes since the page was first entered are listed at the end inside an <added\_elements> block.\\
If the elements you need to interact with do not appear inside any rendered subtree, or no available action makes progress, call view\_all() to re-expose every region's full subtree for the rest of this page.\\
</observation\_space>
\end{tcolorbox}
\caption{Prompt for action selection with \pipeline{}. Only the additions to Figure~\ref{pmt:action} are shown.}
\label{pmt:action_pipeline}
\end{figure}



\end{document}